\documentclass[11pt]{article}

\usepackage[final]{acl}

\usepackage{times}
\usepackage{latexsym}
\usepackage{amsmath}
\usepackage{booktabs}
\usepackage{tabularx}
\usepackage{todonotes}

\newcommand{\cometbase}{COMET$^{22}_{\text{DA}}$}
\newcommand{\cometkiwibase}{COMETKiwi$^{22}_{\text{DA}}$}
\newcommand{\gembaqe}{\mbox{GEMBA$_{\text{ESA}}$-QE}}
\newcommand{\esadataacr}{\mbox{LocHD}}
\newcommand{\locqe}{LocQE}
\newcommand{\locchecklist}{\mbox{LocCheck}}

\usepackage[T1]{fontenc}

\usepackage[utf8]{inputenc}

\usepackage{microtype}

\usepackage{inconsolata}

\usepackage{graphicx}
\usepackage{tikz}
\usepackage[dvipsnames]{xcolor}
\usetikzlibrary{positioning, arrows.meta, shapes, math}
\usepackage{standalone}
\usepackage[inline]{enumitem}

\title{\locqe: Principled Domain Adaptation for Localisation Quality Estimation by Leveraging Post-Edits}

\author{
  \textbf{Kathy Hämmerl\textsuperscript{1,2,}}\thanks{Bulk of work done during an internship at LILT.} \and
  \textbf{Gabriel Bretschner\textsuperscript{3}} \and
  \textbf{Joern Wuebker\textsuperscript{3}} \\
\textsuperscript{1}Technical University of Munich,
 \textsuperscript{2}Munich Center for Machine Learning,
 \textsuperscript{3}LILT\\
 \small{
   \textbf{Correspondence:} \texttt{k.haemmerl@tum.de}
 }
}

\begin{document}
\maketitle
\begin{abstract}
Learned quality estimation (QE) models such as COMETKiwi are widespread and work well for general machine translation evaluation.
However, they are known to struggle on unseen domains, limiting their performance in a real-world localisation context.
We show that they are insensitive to some important factors in localisation, such as whether numbers are translated accurately, or even whether the correct number of spaces and punctuation are preserved in a translation.
Further, a key capability for optimisation of machine translation is the ability of QE models to accurately rank different translations of a single segment, which suffers significantly from the domain transfer.
In the absence of large-scale direct assessment data, we propose principled fine-tuning approaches to reduce the domain gap with even small amounts of post-editing data.
Using a multi-task fine-tuning approach and a simple tokeniser intervention, we create a QE model which proves markedly better at distinguishing preferred post-edits from rejected initial translations in a localisation context.
We show that preferences and artificial continuous scores stabilise each other, and argue that to calibrate metrics both in terms of their absolute scores and comparisons between translation of the same source, both types of signal are needed.
\end{abstract}

\section{Introduction}

Learned neural machine translation metrics are known to struggle in unseen domains \citep{zouhar-etal-2024-fine}.
Both reference-based metrics and quality estimation models such as COMETKiwi \citep{rei-etal-2022-cometkiwi}, have been trained primarily on news and wiki text.
In real-life commercial localisation workflows, the domains vary from spec sheets to user interface labels to legal text \citep[e.g.,][]{buschbeck-exel-2020-parallel,lin-etal-2022-automatic}.
Additionally, localisation workflows require the consistent use of dedicated terminology, and following specific style guidelines.

\begin{figure}
    \centering
    \includegraphics[width=\linewidth]{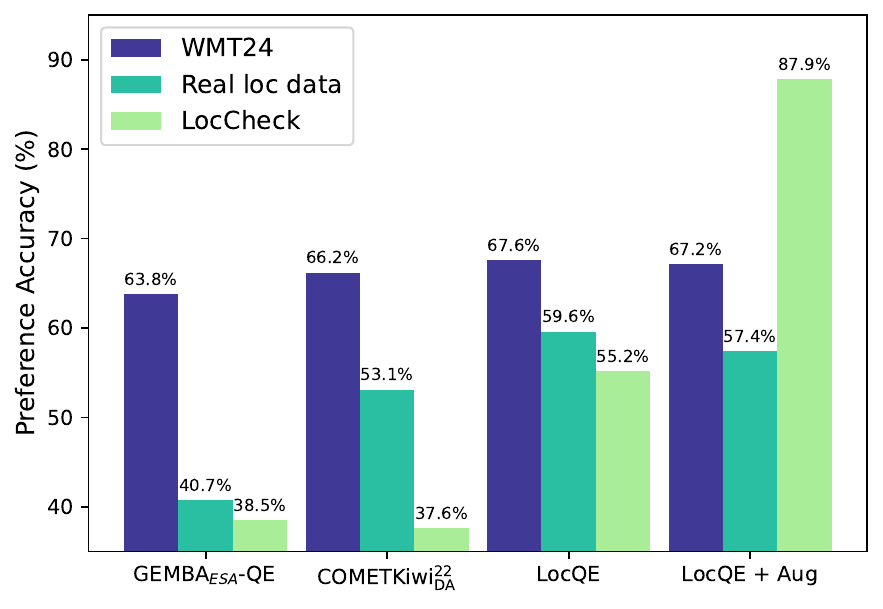}
    \caption{Comparison of \gembaqe, \cometkiwibase, and our \locqe\ models: Preference accuracy  on public data (WMT24) is much higher than on real localisation data (here: internal post-edit data), or on long-tail issues important to localisation (\locchecklist).
    Domain adaptation reduces the gap significantly, but to solve for long-tail issues, data augmentation is needed.
    }
    \label{fig:domain-shift-qe}
\end{figure}

As illustrated in Figure~\ref{fig:domain-shift-qe}, quality estimation (QE) is highly vulnerable to this domain shift, with \gembaqe\ and \cometkiwibase\ both unable to consistently identify the preferred translation on localisation data, while performing well on WMT data.

At the same time, QE is highly relevant for localisation workflows.
Firstly, due to the nature of post-editing, any available references are frequently similar to the hypotheses, biasing reference-based metrics towards the originally used machine translation model.
Secondly, effective QE is needed for finetuning workflows such as direct quality optimisation (DQO; \citealp{uhlig-etal-2025-cross}).
In this context, an especially important capability is \textit{segment-level ranking} of different translations of the same source, so that QE can reliably pick preferred translations.

\begin{table*}[t]
    \centering
    \small
    \begin{tabular}{p{13cm}r}
    \toprule
         & \textbf{\cometkiwibase} \\
    \midrule
        These cargo pants are the easiest to dress when you are attending an outing! \\
        $+$ Ce pantalon cargo est le plus facile à habiller lorsque vous participez à une sortie{\color{green!70!black}[NBSP]}! & 84.83 \\
        $-$ Ce pantalon cargo est le plus facile à habiller lorsque vous participez à une sortie ! & 84.83 \\
    \midrule
        Since they caused the problem, maybe they can do something to \_\_\_\_\_\_ it too. \\
        $+$ Nachdem sie das Problem verursacht haben, können sie vielleicht auch etwas tun, um es zu \_\_\_\_\_\_. & 78.89 \\
        $-$ Nachdem sie das Problem verursacht haben, können sie vielleicht auch etwas tun, um es zu {\color{red} lösen}. & 84.20 \\

    \midrule
        Can you confirm that the input 7777777 has a number with seven digits? &  \\
        $+$ Können Sie bestätigen, dass die Eingabe 7777777 eine Zahl mit sieben Ziffern hat? & 85.91 \\
        $-$ Können Sie bestätigen, dass die Eingabe {\color{red} 777777} eine Zahl mit sieben Ziffern hat? & 85.57 \\
    \bottomrule
    \end{tabular}
    \caption{Exemplars of localisation-specific issues that COMETKiwi misses.
    The model cannot see non-breaking spaces (``[NBSP]''), may prefer a grammatical sentence with a hallucination over the correct translation of a cloze (``\_\_\_\_\_\_''), and is not sensitive enough to errors in long numbers.
    }
    \label{tab:exemplars}
\end{table*}

Not only is there a marked drop in performance compared to WMT data (Figure~\ref{fig:domain-shift-qe}), but we also empirically observe that
\begin{enumerate*}[label=\arabic*)]
  \item Translators and reviewers react strongly to specific issues that the models either cannot see at all---such as the incorrect use of non-breaking spaces---, or even reward, such as hallucinating a word to fill a cloze; and
\item For long numbers, incorrect translations are not consistently distinguished and caught.
\end{enumerate*}
Examples are shown in Table~\ref{tab:exemplars}.
These are mostly long-tail issues, meaning they occur rarely in training or test data, but since human translators often consider them trivially wrong, they may erode trust if not caught.

To address the domain shift, we aim to adapt a QE model in the absence of large-scale human score data by leveraging post-edits.
In addition to this adaptation training, we analyse QE model performance on the aforementioned long-tail issues, and propose adapting the input tokeniser so that models can actually ``see'' some of these issues.

Although other high-quality QE models have been proposed since its release, \cometkiwibase\footnote{We use this notation to disambiguate model versions, referring specifically to \texttt{wmt22-cometkiwi-da}.} represents a very good tradeoff between strong performance and a lightweight model. It is both cheap to run and efficient to fine-tune for quick iteration.
It has also been found to outperform LLM-based metrics at segment-level ranking \citep{moghe-etal-2025-machine}.
Therefore, we use \cometkiwibase\ as the base model for experimentation in this work.

Our key contributions are:
\begin{enumerate}
    \item We formulate a lightweight, data-efficient approach to domain adaptation for QE models without human direct assessments, using post-edits and simple heuristics as our training signal. Tokeniser adaptation allows the model to learn specific relevant patterns.
    \item We describe a new challenge set for localisation-specific issues (\locchecklist), and a new human-annotated localisation dataset (\esadataacr).
    \item We demonstrate improvements on the two newly-proposed datasets, as well as for preference accuracy on natural target-domain data.
\end{enumerate}

Our code for regenerating \locchecklist, and the \esadataacr\ data, are released on GitHub.\footnote{\url{https://github.com/lilt/loc-qe}}

\section{Related Work}

\subsection{COMETKiwi models}
The COMET framework was initially proposed in \citet{rei-etal-2020-comet}.
In it, multilingual encoders are fine-tuned to predict normalised human direct assessments of machine translations.
The first COMET models required a reference,
but later iterations expanded to QE with COMET-QE \citep{rei-etal-2021-references} and COMETKiwi \citep{rei-etal-2022-cometkiwi}, as well as error span prediction with xCOMET models \citep{guerreiro-etal-2024-xcomet}.

\subsection{Domain Adaptation for Metrics}
Similarly to any fine-tuned language model, the COMET models can struggle when applied to unseen domains, as \citet{zouhar-etal-2024-fine} show for reference-based neural models.
\citet{sharami-etal-2023-tailoring} also discuss this phenomenon for quality estimation, and fine-tune a model pre-trained on out-of-domain data with progressively more specific target-domain data.
They produce target-domain augmented data by inferencing an MT model and scoring its hypothesis against the existing reference translation via TER \citep{snover-etal-2006-study}, and evaluate their QE model on correlation with HTER (``Human-targeted Translation Edit Rate'', referring to the edit rate between a hypothesis and a human post-edit given that hypothesis).
The present work instead evaluates both against preferences derived from post-edits and ESA scores (Error Span Annotation, \citealp{kocmi-etal-2024-error}) from professional annotators.
We use two types of training signal derived from these post-edits:
continuous scores and preference pairs.

Another related approach is ``poor man's quality estimation''  \citep{zouhar-etal-2023-poor}.
They suggest pre-training quality estimation with large-scale \textit{metric estimation} (i.e., predicting the scores of a reference-based metric) before fine-tuning with small amounts of human labels.
In the present work, we instead use metric estimation as a continuous training signal for domain adaptation.

MS-COMET \citep{kocmi-etal-2022-ms} is a complete re-training of COMET and COMET-QE with much larger training corpora of 2~M and 3.5~M human-annotated segments, respectively.
They cover over 110 languages and 15 domains, using only annotations from professional translators.
In contrast to the original COMET models, they use raw direct assessment scores, despite the risk of scores becoming incomparable across languages.
Our approach requires much less data, and
we keep z-score normalisation intact.

\citet{schmidt-etal-2026-watches} analyse system-level and segment-level accuracy of metrics across domains, showing that while human annotators exhibit independent noise around a ``true'' rating, metrics tend to exhibit correlated errors.
They aim to disentangle domain shift from noise in human labels, and demonstrate metric biases on out-of-domain data.

\subsection{Contrastive Training}

Contrastive training has been proven effective for representation learning in the past, such as in \mbox{InfoXLM} \citep{chi-etal-2021-infoxlm} or mSimCSE \citep{wang-etal-2022-english}.
Preference tuning of LLMs has enjoyed great popularity in recent years, from RLHF \citep{christiano-etal-2017-deep,ouyang2022traininglanguagemodelsfollow} and direct preference optimisation (DPO; \citealp{rafailov-etal-2023-direct}), to specialised workflows such as DQO \citep{uhlig-etal-2025-cross}.
Like the present work, \citet{berger-etal-2024-post} discuss using machine translated segments and their corresponding post-edits as preference pairs, though they apply them to DPO on LLMs for machine translation.

The original COMET paper includes a ranking-based metric version, which creates a score from the Euclidean distance of the hypothesis to both the source and reference, learned via Triplet Margin Loss \citep{rei-etal-2020-comet}.
However, this method relies on a reference, unlike our approach, and it produces scores directly from model representations rather than using a regression head.
It is also not regularised, meaning that the model tends to distribute scores unevenly.
The COMET-Rank model was dropped in later iterations of COMET.

Recently, \citet{proietti-etal-2026-pear} proposed a relative quality estimation framework that takes in a source segment and two candidate translations, thus focusing on segment-level ranking accuracy.
However, their system cannot return absolute scores, making the ratings more difficult to interpret.

The most conceptually similar work to ours is \citet{tan-monz-2025-remedy}, who use raw human scores as inputs to induce preference pairs with a variable margin, then train using a Bradley-Terry objective.
Similarly to us, they are motivated by relatively poor metric performance at a segment level.
Compared to our work, they train larger models using a reference and WMT-domain data, whereas we focus on domain adaptation and do not have access to an independent reference or continuous human ratings.
By leveraging post-edits into both preferences and continuous scores, using a multi-task training setup, we avoid the need for post-hoc calibration of output scores.

\subsection{Metric Challenge Sets}
ACES and SPAN-ACES \citep{amrhein-etal-2023-aces, moghe-etal-2025-machine} are metric challenge sets intended to profile strengths and weaknesses of machine translation metrics on different sub-categories of translation accuracy issues.
They do not cover surface issues other than some relatively easy punctuation errors.
ACES covers many language pairs, though with a very small number of examples for some.
Metrics are scored on whether a ``good'' translation is ranked higher than a translation with the error under examination.
We use a similar design for our \locchecklist\ dataset, but focus on localisation-specific issues such as those discussed in the introduction.

\section{Datasets}\label{sec:datasets}

In this section, we summarise the datasets used in this study.
Table~\ref{tab:data-sizes} lists the dataset sizes.

\begin{table}[t]
    \centering
    \begin{tabular}{lrrr}
    \toprule
        \textbf{Corpus} & \textbf{train} & \textbf{dev} & \textbf{test} \\
        \midrule
        ACED corpus & 17,455 & 7,884 & 9,922 \\
        ACED (prefs) & 2,179 & 1,056 & 959 \\
        Internal & 10,000 & 1,000 & Table~\ref{tab:dataset-sizes-internal} \\
        \esadataacr\ & -- & -- & 3,000 \\
        WMT24 (prefs) & -- & -- & 5,718 \\
        ACES & -- & -- & 36,476 \\
        \bottomrule
    \end{tabular}
    \caption{Dataset sizes, in number of rows, used in our experiments. For details on the per-language distribution and number of test segments we create from internal data, see Appendix Table~\ref{tab:dataset-sizes-internal}.
    }
    \label{tab:data-sizes}
\end{table}

\paragraph{Internal Localisation Data.}

Internally, we have access to a large collection of localisation-specific data across various client domains and language pairs, of which a large percentage has been post-edited by humans.
The data includes, for example, technical documentation, user interface text, lists of features, and other specialised content.
There is a history of edits for each segment, but there are no at-scale direct assessments.

In order to adapt QE models to the localisation domain, we process the post-edits in two different ways, producing both artificial continuous scores and preference pairs.
To obtain preference pairs, we take machine translated segments as the negative sample, and the post-edited version provided by a human reviewer as the positive sample, provided an edit was made.
We have no independent reference in this setting.
To create artificial scores, we treat the human-edited segment as a reference and use a surface metric (chrF++) to provide a score for the machine-translated segment.
Additionally, we use heuristics to identify major errors that resulted in small chrF changes, and reduce the score in such cases.
We then apply z-score normalisation, treating each client project as one annotator, and rescale the scores to the range [0,1].
Further data processing details can be found in Appendix~\ref{app:sec-data-proc}.
Additionally, in Appendix~\ref{subsec:quality-val}, we describe a small validation study for both types of internal data.

Despite the potential subjectivity of post-edits \citep[cf.][]{popovic-2021-agree}, they are the best human-created quality signal we have.
Therefore, we use preference accuracy as the main evaluation metric on this data.

\paragraph{ACED Localisation Data.}

The ACED corpus is a public collection of English-to-German localisation data, initially for Translation Error Correction \citep{lin-etal-2022-automatic}.
For each segment, it includes a source, target, and perturbed target, where the perturbed target represents the initial hypothesis and the target is the post-edited segment.
The data is spread across three sub-datasets with different domains: \textsc{Asics} (marketing copy for an activewear company), \textsc{Emerson} (industrial product listings for a manufacturer), and \textsc{DigitalOcean} (software engineering tutorials).

We use the predefined train, dev, and test splits, but merge them across sub-datasets for a larger data size.
We apply the same transformations as for our internal data to obtain preference pairs and continuous scores.
Note that the number of preference pairs is especially limited, since not all segments in ACED received a post-edit.
Table~\ref{tab:data-sizes} lists the merged data sizes for ACED.

\paragraph{Localisation Data with Human Scores (\esadataacr).}

For evaluation purposes, we construct a \textbf{Loc}alisation \textbf{H}uman \textbf{D}ata corpus.
We translate 300 segments from English into ten target languages\footnote{Arabic, German, Spanish, Finnish, French, Japanese, Polish, Portuguese, Swedish, and Turkish.} using a selection of different commercial models, then collect Error Span Annotations \citep{kocmi-etal-2024-error} from professional translators, who were paid standard rates.
We refer to this corpus as \esadataacr\ from here on.

The source data originates from a real client project, and includes primarily user interface text and help documentation.
Since this data does not contain personally identifiable information in the translation segments, we only pseudonymise the annotators involved before release.
There were between three and five annotators involved in the project per target language.
Annotations were reviewed on a spot-checking basis, but independent overlapping annotations were not collected.
\esadataacr\ contains hypotheses but no references.

We compute z-scores per annotator and evaluate QE models on a global correlation (Kendall's~$\tau$) with the z-scores.
This data is not suitable to transform into preference pairs because we only have one scored translation per segment, so we only use it to compute correlations.

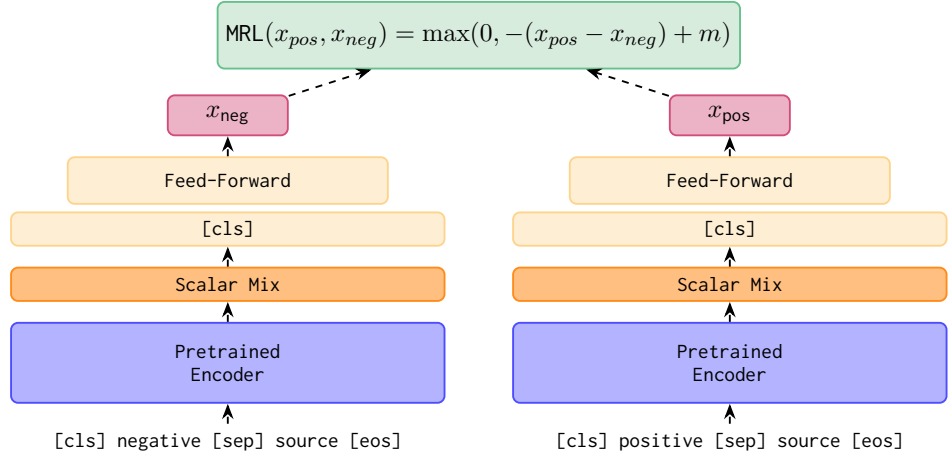
\begin{figure*}[t]
\centering
\begin{tikzpicture}[
    encoder/.style={rectangle, rounded corners=3pt, fill=blue!30, draw=blue!70, thick, minimum width=6.5cm, minimum height=1.2cm, text centered, align=center, font=\ttfamily\small},
    pooling/.style={rectangle, rounded corners=3pt, fill=orange!50, draw=orange!90, thick, minimum width=6.5cm, minimum height=0.5cm, text centered, align=center, font=\ttfamily\small},
    concat/.style={rectangle, rounded corners=3pt, fill=Dandelion!20, draw=Dandelion!60, thick, minimum width=6.5cm, minimum height=0.5cm, text centered, align=center, font=\ttfamily\small},
    feedforward/.style={rectangle, rounded corners=3pt, fill=Dandelion!20, draw=Dandelion!60, thick, minimum width=4.8cm, minimum height=0.7cm, text centered, align=center, font=\ttfamily\small},
    output/.style={rectangle, rounded corners=3pt, fill=purple!30, draw=purple!70, thick, minimum width=1.8cm, minimum height=0.6cm, text centered, align=center, font=\ttfamily},
    loss/.style={rectangle, rounded corners=3pt, fill=Green!15, draw=Green!60, thick, minimum width=3cm, minimum height=1cm, text centered, align=center, font=\ttfamily},
    input/.style={font=\ttfamily\small, text=black},
    branchlabel/.style={font=\small\bfseries, text=black},
    arrow/.style={->, >=Stealth, thick, draw=black, dashed},
    arrowup/.style={->, >=Stealth, thick, draw=black, dashed},
    node distance=1cm
]

    \coordinate (center) at (0,0);
    
    \node[encoder, left=0.5cm of center] (encneg) {Pretrained\\Encoder};
    \node[pooling, above=0.3cm of encneg] (poolneg) {Scalar Mix};
    \node[concat, above=0.3cm of poolneg] (concatneg) {[cls]};
    \node[feedforward, above=0.1cm of concatneg] (ffneg) {Feed-Forward};
    \node[output, above=0.3cm of ffneg] (xneg) {$x_{\text{neg}}$};
    
    \node[encoder, right=0.5cm of center] (encpos) {Pretrained\\Encoder};
    \node[pooling, above=0.3cm of encpos] (poolpos) {Scalar Mix};
    \node[concat, above=0.3cm of poolpos] (concatpos) {[cls]};
    \node[feedforward, above=0.1cm of concatpos] (ffpos) {Feed-Forward};
    \node[output, above=0.3cm of ffpos] (xpos) {$x_{\text{pos}}$};

    \node[loss, above=4.4cm of center] (loss) {$\text{MRL}(x_{pos}, x_{neg}) = \max(0, -(x_{pos} - x_{neg}) + m)$};

    \node[input, below=0.3cm of encneg] (neginput) {[cls] negative [sep] source [eos]};
    
    \node[input, below=0.3cm of encpos] (posinput) {[cls] positive [sep] source [eos]};

    \draw[arrow] (neginput) -- (encneg);
    \draw[arrow] (encneg) -- (poolneg);
    \draw[arrow] (poolneg) -- (concatneg);
    \draw[arrow] (ffneg) -- (xneg);

    \draw[arrow] (posinput) -- (encpos);
    \draw[arrow] (encpos) -- (poolpos);
    \draw[arrow] (poolpos) -- (concatpos);
    \draw[arrow] (ffpos) -- (xpos);

    \draw[arrowup] (xneg) -- (loss);
    \draw[arrowup] (xpos) -- (loss);

\end{tikzpicture}
\caption{Margin Ranking Loss applied to a COMETKiwi model. Figure adapted from \citet{rei-etal-2020-comet,rei-etal-2022-cometkiwi}.}
\label{fig:margin_ranking_loss}
\end{figure*}

\paragraph{\locchecklist.}
Motivated by the observations stated in the introduction and exemplified in Table~\ref{tab:exemplars}, we create a localisation-specific challenge set addressing a specific set of potential issues.
We think of this dataset as a kind of CheckList \citep{ribeiro-etal-2020-beyond}, covering behaviours that may appear simple on the surface but could get lost while optimising for other skills.
Specific points we address here include the translation of long number sequences, the preservation of leading or trailing whitespaces, the correct use of non-breaking spaces in French, the translation of all-caps sentences, and more.

We create both an internal and a public data split:
For the public split, we
augment data from WMT25 and BOUQuET~\citep{teamBOUQuETDatasetBenchmark2025}, whereas the internal split is sourced from real samples in our internal data.
We create an internal split because these specific issues are rarer in public MT datasets.
We include eleven target languages in \locchecklist, all translating from English: Arabic, German, Spanish, Finnish, French, Hindi, Japanese, Polish, Russian, Thai, and Chinese.
Our public split is missing Finnish, Polish and Thai as target languages because they were not present in WMT25 or BOUQuET.

Each row consists of a source, a reference, a positive sample,
and a negative sample which is identical to the positive sample except for the variable under study.
We ensure that the reference differs from the positive and negative samples.
Appendix~\ref{app:loc-checklist-comp} gives a detailed description of each heuristic, and shows the distribution of samples in both splits.
The code for the public test set is released for reproducibility.

\paragraph{Training data augmentation.}
By definition, many phenomena in \locchecklist\ are long-tail issues:
Although they are rated as important by translators, they occur rarely in real-world data.
Therefore, we create augmented training data to ensure the model sees these patterns.
We apply the same transformations that were used to create \locchecklist\ to a portion of our internal data, yielding 5k pairs.

\paragraph{WMT Data.}

COMETKiwi \citep{rei-etal-2022-cometkiwi} was originally trained with normalised direct assessments from the WMT years 2017-20 \citep{bojar-etal-2017-findings, bojar-etal-2018-findings, barrault-etal-2019-findings, barrault-etal-2020-findings}.
We experimented with mixing in data from those years during training to mitigate forgetting, and test on preference pairs derived from WMT24 \citep{kocmi-etal-2024-findings} to validate the general performance of our finetuned metrics.
See Appendix~\ref{app:wmt-data-proc} for how we process this data.

\section{Metric Adaptation}\label{sec:main-method}

\subsection{Multi-Task Training}

Rather than fine-tuning solely on preference pairs or continuous scores, we implement a multi-task training regime.
We randomly mix samples from both datasets---continuous or contrastive---in every single batch.
The respective losses for each type of data are described in this section.
We compute the appropriate loss only for the corresponding samples, making the final batch loss a mixture of both losses.
We balance the influence of each loss by varying the data distribution, empirically selecting a ratio of 2:1 in favour of continuous scores (cf. ablation in Appendix~\ref{app:data-mix-ablation}).

\paragraph{Contrastive training with preference pairs.}
For training with preference pairs, we use the Margin Ranking Loss (MRL; \citealp{herbrich-etal-2000-large}) to encourage the model to score the preferred hypothesis above the rejected one:

\begin{multline}
    \text{MRL}(x_{pos}, x_{neg}) = \\ \max(0, -(x_{pos} - x_{neg}) + m)
\end{multline}

where $x_{pos}, x_{neg}$ are the scores of the preferred and rejected inputs, respectively, and the margin $m$ is a hyperparameter.
Figure~\ref{fig:margin_ranking_loss} shows the loss schematic for MRL applied to COMETKiwi.

\paragraph{Regression on continuous scores.}

With the artificial continuous scores, we use the standard mean-squared error (MSE) loss.
A single score is computed for the hypothesis and compared against the target score described in~\S~\ref{sec:datasets}.

\subsection{Tokeniser Adjustments}

Some of the \locchecklist\ rules involve special characters such as non-breaking spaces, which \cometkiwibase\ ignores due to its tokeniser.
The process of tokenisation involves multiple steps, all of which influence the outcome \citep[cf.][]{schmidt-etal-2024-tokenization}:
Normalisation, pretokenisation, tokeniser inference, and
post-processing.
We propose adjusting the pretokenisation step of the encoder model's tokeniser, adding new tokens for special whitespace and other control characters.
These are targeted interventions which somewhat change the distribution of tokenised texts seen by the model, but leave the majority of segments unchanged.
Specifically, we keep trailing and multiple spaces, and add missing tokens which enable the model to ``see'' the most relevant special characters.
Details of the tokeniser modifications are in Appendix~\ref{app:tok-details}.

\subsection{Models}

\paragraph{Main system.}
We finetune \cometkiwibase\ with the updated tokeniser using the multi-task loss on internal data.
We use 10k samples of MSE data and 5k preference pairs for MRL.
For comparison, we show runs with each loss individually, using a total of 15k samples respectively.
Finally, we add the 5k samples of augmented preference pairs for a \locqe\ version with data augmentation.
Since these augmented data are minimal pairs exemplifying the rules in \locchecklist, they only provide an explicit signal to the model in cases where the baseline ranking is wrong, and none otherwise.

\paragraph{Public data analog.}
Similarly, we finetune models on the ACED data, using all available preference pairs and inferred continuous scores for the respective losses.
Due to the limited training set size in ACED, we also added 5.7k WMT data samples to both of these training sets, keeping all data for the multi-task training here.

\paragraph{Hyperparameters.}
The margin $m$ for the MRL loss is set to $0.02$ (cf. ablation in Appendix~\ref{app:hyperparams}).
We unfreeze the entire encoder, including the embedding layer, from the beginning of finetuning, in order to train the added token embeddings.
Instead of freezing the encoder, we add learning rate warmup to stabilise training, and increase both the effective batch size and the learning rates.
We use 50 warmup steps, an encoder learning rate of 1e-5, and a feedforward layer learning rate of 1e-4.
The effective batch size is 512.
We apply early stopping with a checkpoint every epoch, training for up to five epochs.
The results show individual runs with a fixed random seed.

\paragraph{GEMBA Baseline.}

We additionally compute reference-free \gembaqe\ \citep{kocmi-federmann-2023-large, kocmi-etal-2024-error} with GPT-4.1 as the backbone.
GEMBA has been shown to yield good system-level pairwise accuracy but perform relatively poorly at the segment level.

\section{Domain Adaptation Results}

\begin{figure*}[t]
    \centering
    \includegraphics[width=0.48\linewidth]{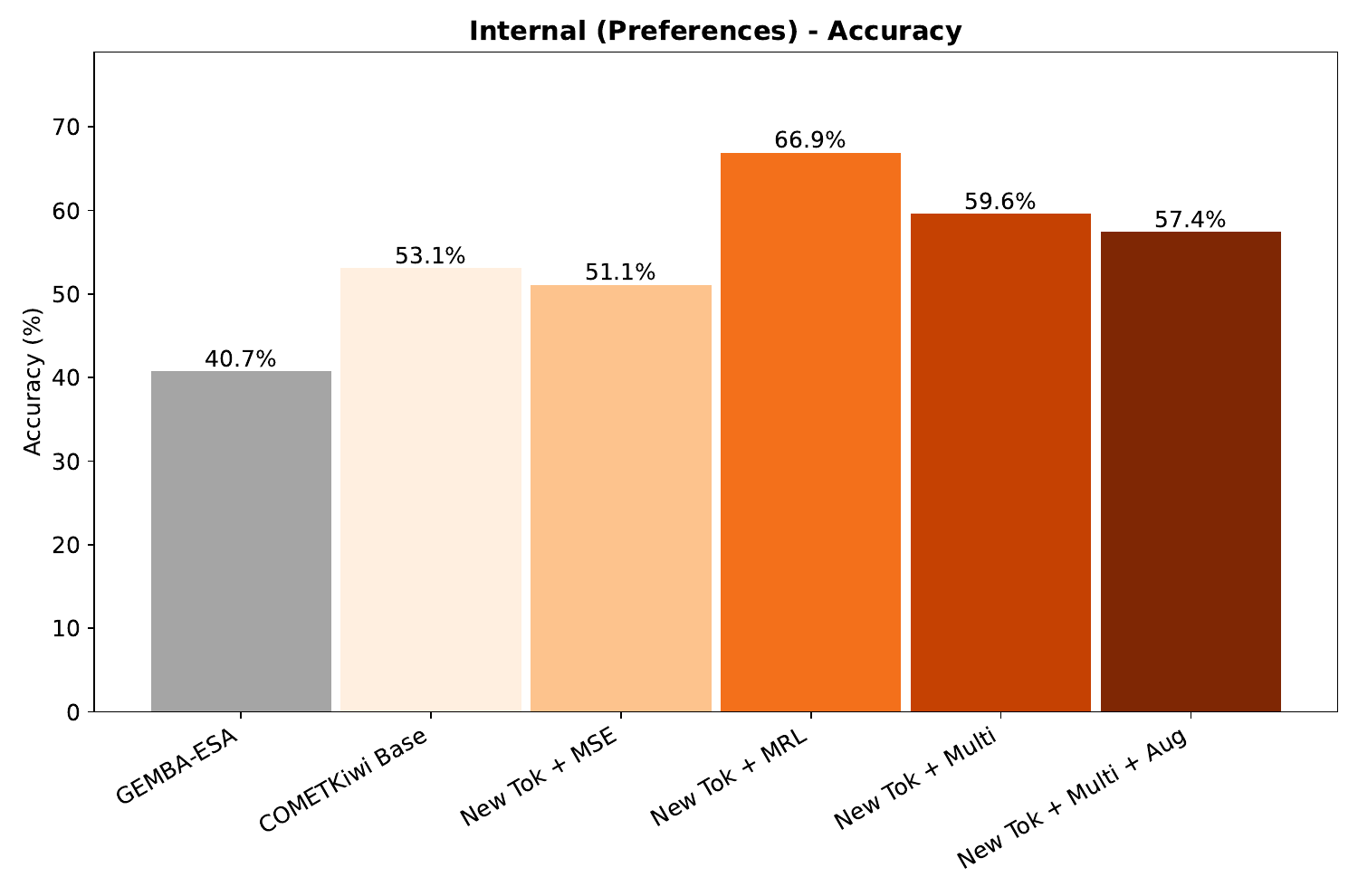}
    \includegraphics[width=0.48\linewidth]{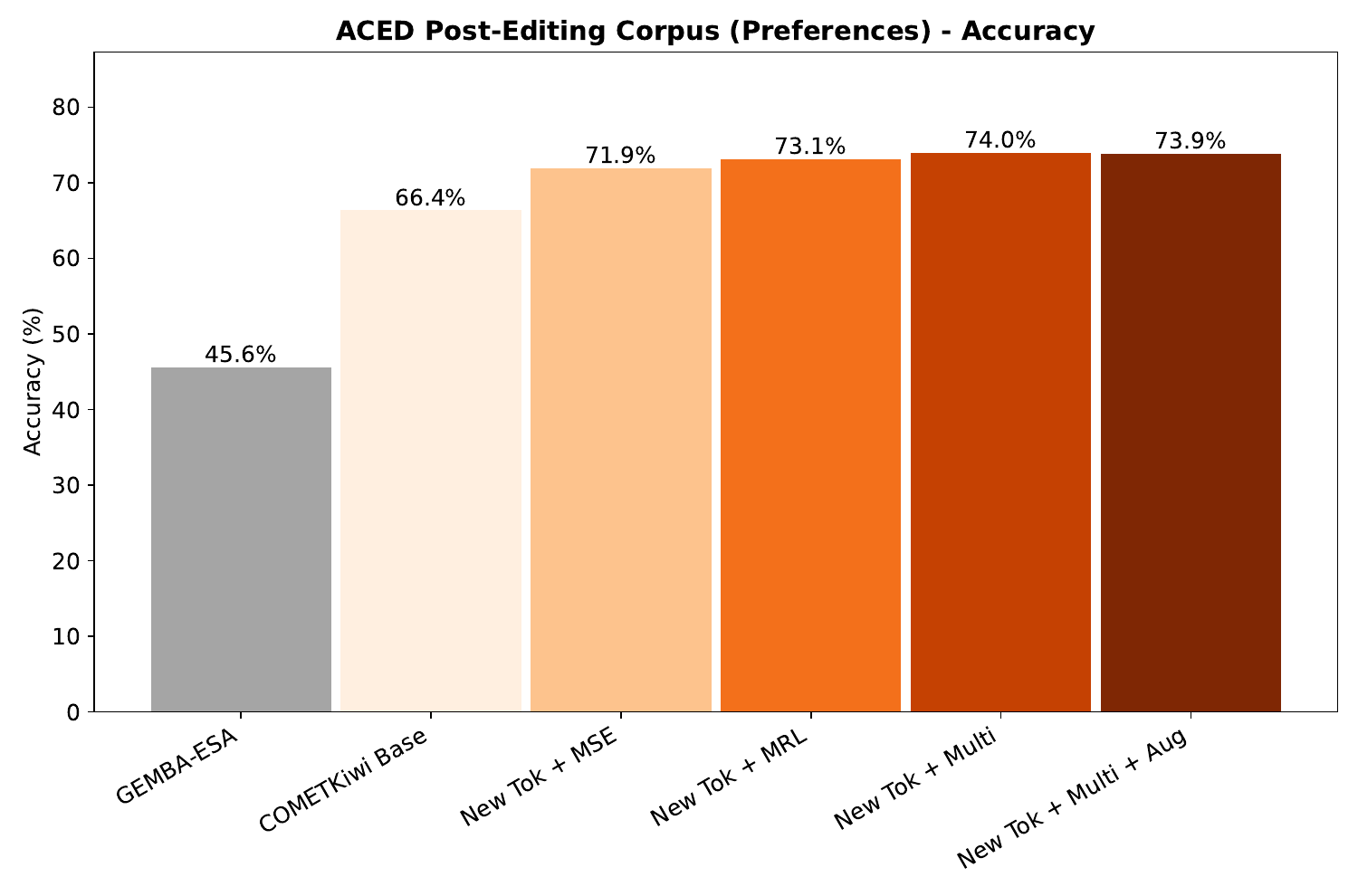}
    \caption{Performance on the respective target test set, in terms of preference accuracy. Left: Internal data. Right: ACED corpus.
    Per-language-pair performance for the internal data is listed in Appendix Table~\ref{tab:results-pref_acc-segment-history-test-prefs}.
    }
    \label{fig:target-corpus-acc}
\end{figure*}

\begin{figure}[t]
    \centering
    \includegraphics[width=\linewidth]{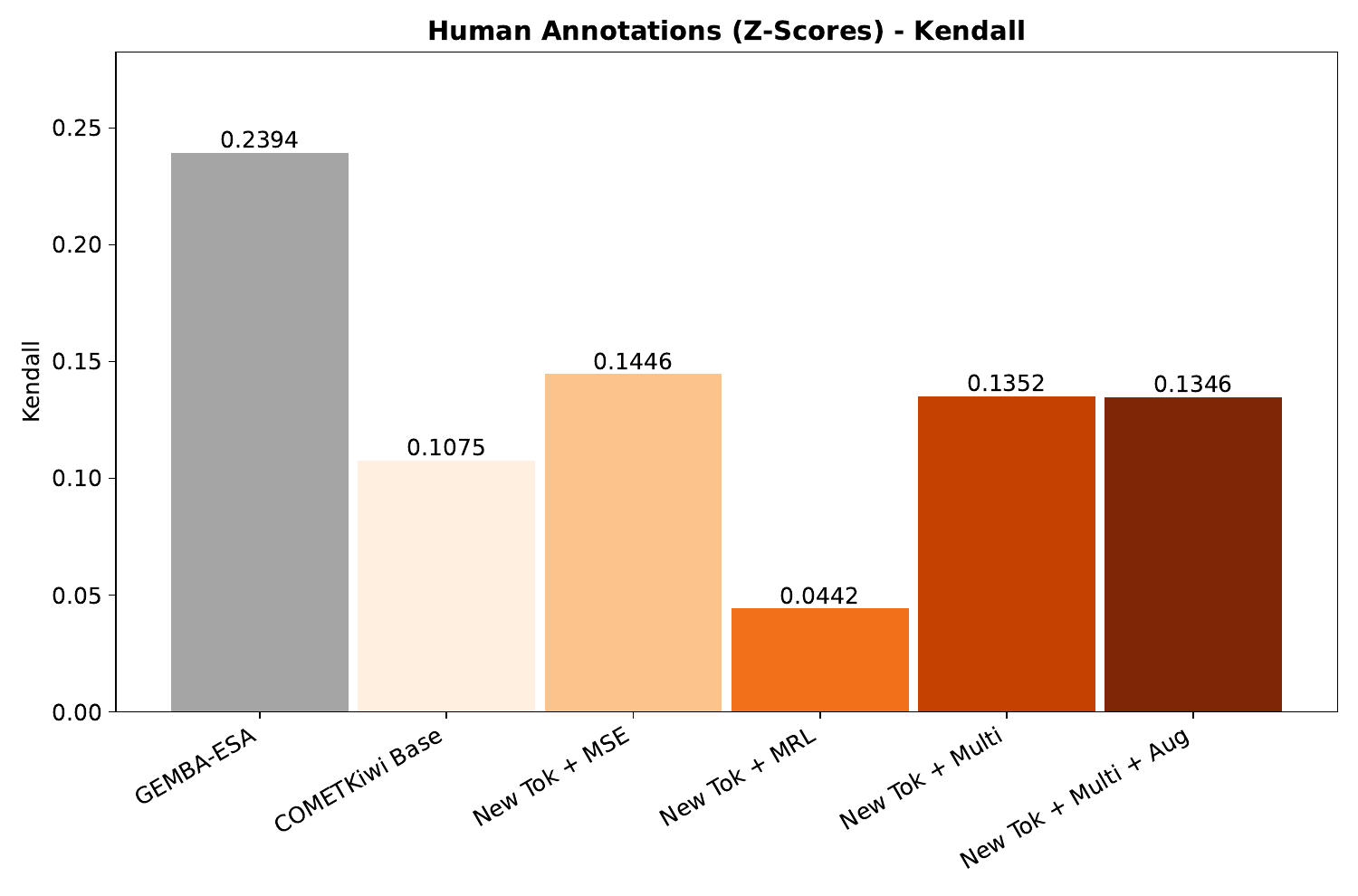}
    \caption{Overall performance on \esadataacr, in terms of global Kendall's $\tau$. Fine-tuning was done on internal data.
    Per-language-pair performance is listed in Appendix Table~\ref{tab:results-kendall_corr-snap_internal}.
    }
    \label{fig:snap-kendall-overall}
\end{figure}

\begin{figure*}[tb]
    \centering
    \includegraphics[width=0.49\linewidth]{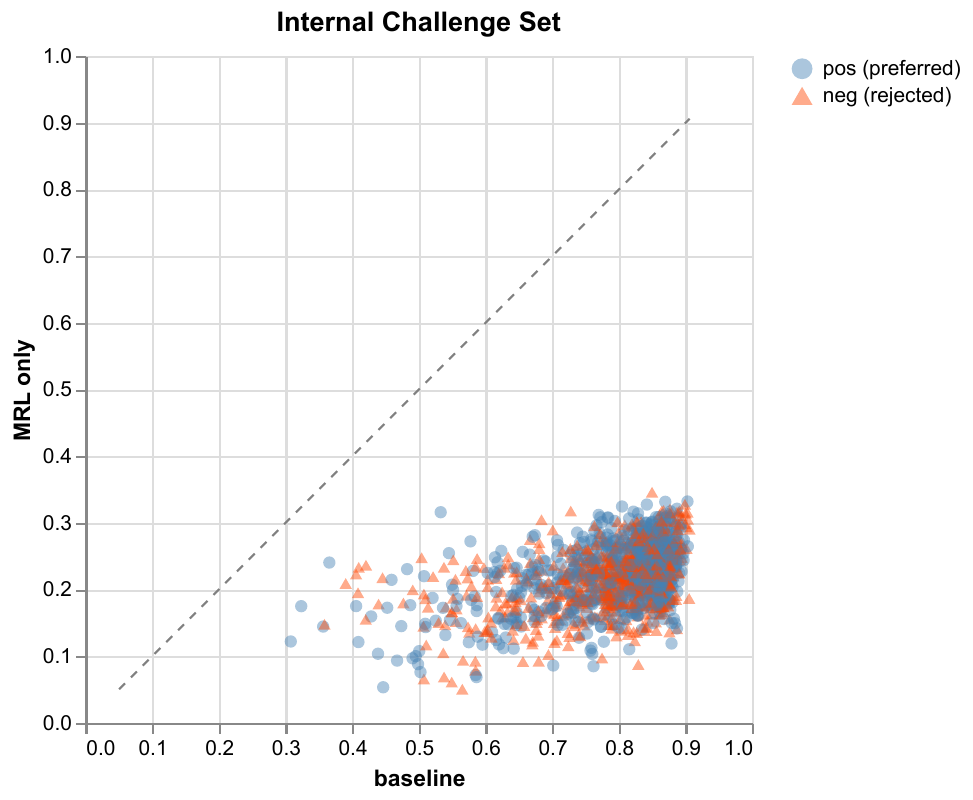}
    \includegraphics[width=0.49\linewidth]{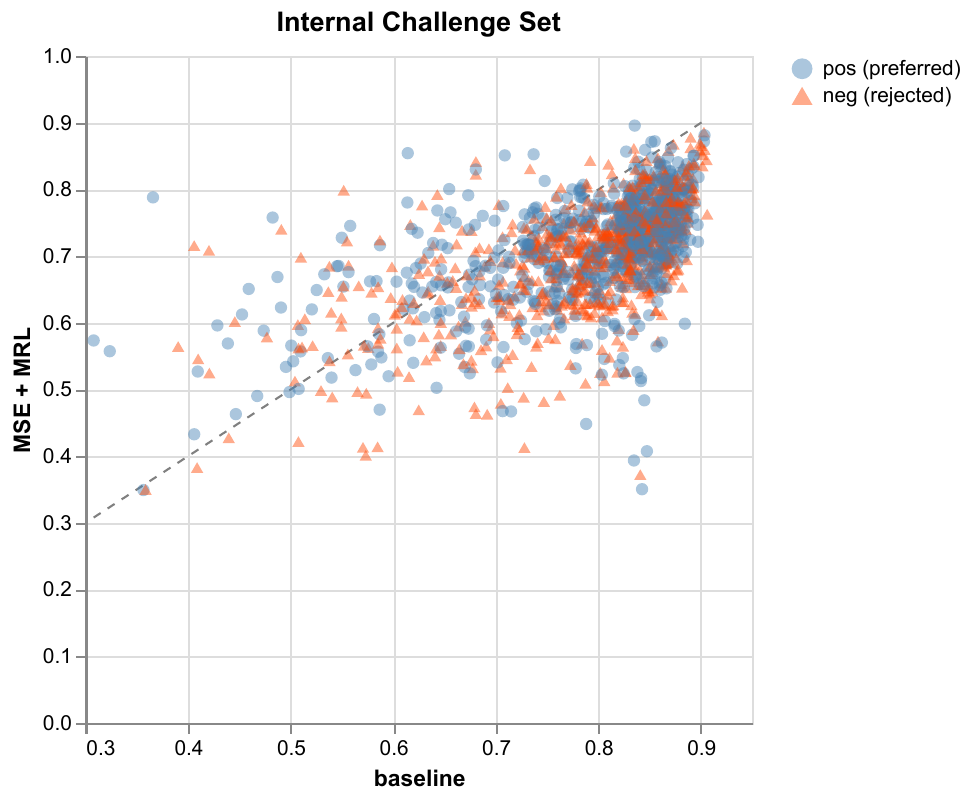}
    \caption{Scatter plots contrasting fine-tuned vs. baseline COMETKiwi scores on internal \locchecklist\ split.
    Left: Model fine-tuned only with MRL; Right: Multi-task training.
    }
    \label{fig:challenge-set-distribution-collapse}
\end{figure*}

\subsection{Effective domain adaptation leveraging a small number of post-edits}
Figure~\ref{fig:target-corpus-acc} shows the accuracy of the QE models at selecting the preferred translation over the rejected on the respective target test set.
The baseline \cometkiwibase\ struggles to identify preferred segments in our internal data, while it performs better initially on the ACED corpus.
\gembaqe\ performs even worse at this task.
Both the contrastive and the multi-task training consistently improve preference accuracy on the target corpus, adding between 6.1 and 13.8 percentage points.
MSE fine-tuning worsens preference accuracy on internal data, while for ACED, the addition of original training data seems to reduce forgetting and improve overall performance.

In Appendix Table~\ref{tab:results-gen-cap}, we show performance of the fine-tuned models on WMT24 preference accuracy and ACES score as a proxy of general capability.
We find that WMT24 preference accuracy stays within a relatively small range regardless of fine-tuning objective, though MRL-only with internal data degrades it slightly and multi-task training improves it somewhat.
ACES score collapses with MRL-only training on internal data, and suffers almost regardless of the setting.
When training with MRL on ACED plus original training data, ACES score is relatively unaffected.
With the exception of MRL-only on internal data, the fine-tuned models also still outperform \gembaqe\ on ACES.
We surmise that including original training data helps to stabilise general model capabilities, but so does the multi-task training setup.
Our analysis in~\S~\ref{subsec:analysis} sheds further light on this observation.

Meanwhile, Figure~\ref{fig:snap-kendall-overall} shows global Kendall's $\tau$ on the \esadataacr\ corpus.
\cometkiwibase\ initially performs much worse at this than \gembaqe, improving somewhat with our internal data training.
Notably, MSE training improves global Kendall the most, while MRL-only training further worsens performance on global Kendall despite the matching domain.
Multitask training strikes a balance between the two, yielding an overall improvement on both preference accuracy and global Kendall.
Next, we analyse why this is the case.

\subsection{Score distribution analysis}\label{subsec:analysis}

We observe that global correlation and segment-level accuracy conflict, with models that perform well by one of these measures tending to do worse on the other.
This is in keeping with analysis by \citet{diianni-deutsch-2025-dont}, who emphasise the difference between Global and Segment-Wise correlations: Global correlations try to place translations on an overall distribution of human scores, while segment-wise correlations or accuracy compare translations given the same source.
Importantly, performance on one of these aspects does not translate directly to the other---for instance, human score noise in a regression setting \citep[cf.][]{schmidt-etal-2026-watches} might obscure direct comparisons.
We argue that both are relevant in a localisation context, but segment-wise performance is more well-defined in our setting.

The two losses, MSE and MRL, intuitively represent a \textit{global} and a \textit{segment-wise} objective.
When training purely on continuous scores, the model learns from isolated scores given to translations of many different sources, and can merely approximate comparisons between different translations of the same source.
By contrast, when training purely on preference pairs, the model learns to compare translation quality given the same source, but has no anchor for the absolute score value.

To visualise the issue, we plot scores on the internal \locchecklist\ split before and after fine-tuning.
Figure~\ref{fig:challenge-set-distribution-collapse} (left) shows MRL by itself:
Since many preference pairs need to be reversed compared to the base model's judgements---remember that baseline performance on internal preference data was just above guessing level---,
the model learns to fulfill the loss by compressing the output scores to a band roughly twice the size of the margin.
This collapsed score distribution then makes it very difficult for the model to preserve ranking across different sources, affecting global Kendall.

\paragraph{Regularising MRL.}
Since our preference-pair data does not contain human scores, or held-out references with which to compute artificial scores, we initially attempt to regularise MRL against \cometkiwibase's original score.
The equation and ablation study are described in Appendix~\ref{app:hyperparams}.
This regulariser partially mitigated score collapse.
However, due to the large number of samples where the baseline is wrong, MRL and the regulariser frequently conflict, producing worse overall outcomes with a stronger regularisation weight.

In our main multi-task training setup, we instead process the data in two separate ways, inducing a chrF++-based score to regress towards for MSE, and creating preference pairs for MRL.
Figure~\ref{fig:challenge-set-distribution-collapse} (right) shows the resulting score distribution on \locchecklist:
Scores remain much more spread out across the range, while still producing better separation between the positive and negative samples.

\section{\locchecklist\ Results}

\begin{table}[t]
\begin{tabularx}{\columnwidth}{@{}Xrr@{}}
\toprule
Model & Internal & Public \\
\midrule
baseline & 45.1 & 30.1 \\
+ new tokeniser & 49.5 & 35.8 \\
\midrule
\multicolumn{3}{@{}l}{\textbf{+ MSE}} \\
\quad base tokeniser & 56.3 & 40.5 \\
\quad new tokeniser & 61.6 & 41.4 \\
\midrule
\multicolumn{3}{@{}l}{\textbf{+ MRL}} \\
\quad base tokeniser & 41.9 & 50.3 \\
\quad new tokeniser & 54.0 & 54.2 \\
\midrule
\multicolumn{3}{@{}l}{\textbf{+ MSE + MRL}} \\
\quad new tokeniser & 64.1 & 46.3 \\
\quad + data augmentation & \textbf{89.6} & \textbf{86.1} \\
\bottomrule
\end{tabularx}
\caption{Effect of the tokeniser and data augmentations on preference accuracy (\%) over \locchecklist. \textbf{Internal} is our in-house challenge set; \textbf{Public} is the same set of heuristics applied to WMT25 system outputs and the BOUQuET test split. Best result per column in \textbf{bold}.}
\label{tab:challenge-set-tokeniser-effect}
\end{table}

Table~\ref{tab:challenge-set-tokeniser-effect} shows results on \locchecklist, both the public and the internal split.
The baseline \cometkiwibase\ performs poorly on both splits, well below guessing on the public split.
We performed additional checks to rule out corrupt data, then tested our fine-tuned models on both splits.
We ablate our tokeniser adjustments and fine-tuning approaches separately from each other.
In Figure~\ref{fig:domain-shift-qe} we show the macro average over both splits.

\paragraph{Tokeniser adaptation helps with learning specific patterns.}

Results on \locchecklist\ demonstrate a large improvement (4-10 percentage points in most cases) simply due to the tokenizer adjustments:
Replacing the tokeniser allows the model to `see' leading and trailing spaces for the first time, for instance.
The All Caps rule is learned better with the new tokeniser, while the NBSP rule can only be learned correctly with the new tokeniser.
Appendix~\ref{subsec:challenge-set-results} offers more detail on how the individual categories respond to the new tokeniser, as well as to the different fine-tuning approaches.
However, the overall results are not satisfactory, even with multi-task fine-tuning:
While accuracy on the internal split has reached 64\%, accuracy on the public split is still below guessing level.

\paragraph{\locchecklist\ needs data augmentation.}

As expected, adding this augmented data allows the fine-tuned model to rank the majority of samples correctly, now approaching 90\% on both data splits.
However, preference accuracy on the main test set drops somewhat across language pairs, compared to the regular multi-task training (see Appendix Table~\ref{tab:results-pref_acc-segment-history-test-prefs}).
This reflects the shift in the training distribution, and indicates that the mentioned ``rules''  may in fact not be consistently applied by the post-editors.
This result suggests that although QE models can be efficiently taught to respect explicit editing rules, heuristics may equally remain a useful tool in the localisation pipeline.

\section{Conclusions}

Real-life localisation data is significantly different from public machine translation and metric benchmarks, introducing a domain shift that affects QE models strongly.
To help with the evaluation of QE models for a localisation context,
we have proposed two new datasets: \locchecklist, focused on locali\-sation-specific long-tail issues, and \esadataacr, a dataset of realistic translations annotated with ESA scores from professional translators.

To bridge the domain gap in the absence of traditional metric training data,
we have demonstrated a data processing approach to leverage in-domain post-edits into two types of signal:
Between-source regression of continuous scores and within-source ordering of preference pairs.
Both approaches individually have their limitations, and indeed, initial results illustrate the trade-off between global between-source correlation and within-source preference accuracy.
To address this trade-off, we combine both losses in a multi-task training approach, where they stabilise each other and yield better overall results.
Our recommendation for future adaptation and/or training of metrics, therefore, is to consistently incorporate training signals at both the global and the segment level.

Using these two signals enables us to perform data-efficient domain adaptation without requiring human scores.
Specifically, we adapted \cometkiwibase\ to localisation data with training sets of only 10k-20k samples, less than 1\% of its original training data.
Along with targeted tokeniser interventions, our fine-tuning approaches yield \locqe\ models more suited to localisation data.
To contend with long-tail issues described by translators, we find that targeted data augmentation is required.

\paragraph{Future Work.}

The application of modified QE models to MT preference optimisation training, such as DQO, could yield deeper insights into their behaviour.
We leave this exploration to future work.

\section*{Limitations}

We acknowledge several limitations of this work.
Primarily, some of the data for our main experiments is internal and cannot be published.
However, we have made an effort to provide comparable experiments on similar, previously published data, showing qualitatively similar results, to share code for reproducibility, and to describe the internal data in detail.

The post-edits used are potentially noisy, particularly in the sense that different stylistic preferences may apply for different client projects.
Our analysis in Appendix~\ref{subsec:quality-val} gives a further sense of the subjectivity of edits, indicating that not every preference pair is likely to yield a helpful signal.
Nevertheless, our results show promising results for domain adaptation even with this limitation.

Additionally, we have demonstrated our approach on only one type of quality estimation model.
This was done for efficiency and cost reasons, as COMETKiwi is a lightweight model that makes it easier to inference and fine-tune.
Tuning a significantly larger model such as MetricX-24 would have been out of scope for this project.

\section*{AI Assistant Information}

Coding assistants were used in implementing the data processing, fine-tuning approaches, and creating the plots.
The outputs were thoroughly checked and tested for purpose.
No LLMs were used in the writing of this article.

\section*{Acknowledgments}

The data processing, tokeniser adaptation, and fine-tuning setups were implemented by KH during an internship at LILT, advised by JW. GB provided significant help in running further ablations and combining the two losses into the multi-task fine-tuning.
KH wrote the paper draft with input from both co-authors.

Thank you to Kaden Uhlig for helpful discussions during the development of this project.
We also thank the anonymous reviewers for their insights and feedback.

This work was co-funded by the European Union (ERC, EPICAL, 101141712). Views and opinions expressed are however those of the author(s) only and do not necessarily reflect those of the European Union or the European Research Council. Neither the European Union nor the granting authority can be held responsible for them.

\bibliography{custom, anthology-1, anthology-2}

\begin{thebibliography}{41}
\providecommand{\natexlab}[1]{#1}

\bibitem[{Amrhein et~al.(2023)Amrhein, Moghe, and Guillou}]{amrhein-etal-2023-aces}
Chantal Amrhein, Nikita Moghe, and Liane Guillou. 2023.
\newblock \href {https://doi.org/10.18653/v1/2023.wmt-1.57} {{ACES}: Translation accuracy challenge sets at {WMT} 2023}.
\newblock In \emph{Proceedings of the Eighth Conference on Machine Translation}, pages 695--712, Singapore. Association for Computational Linguistics.

\bibitem[{Barrault et~al.(2020)Barrault, Biesialska, Bojar, Costa-juss{\`a}, Federmann, Graham, Grundkiewicz, Haddow, Huck, Joanis, Kocmi, Koehn, Lo, Ljube{\v{s}}i{\'c}, Monz, Morishita, Nagata, Nakazawa, Pal, Post, and Zampieri}]{barrault-etal-2020-findings}
Lo{\"i}c Barrault, Magdalena Biesialska, Ond{\v{r}}ej Bojar, Marta~R. Costa-juss{\`a}, Christian Federmann, Yvette Graham, Roman Grundkiewicz, Barry Haddow, Matthias Huck, Eric Joanis, Tom Kocmi, Philipp Koehn, Chi-kiu Lo, Nikola Ljube{\v{s}}i{\'c}, Christof Monz, Makoto Morishita, Masaaki Nagata, Toshiaki Nakazawa, Santanu Pal, and 2 others. 2020.
\newblock \href {https://doi.org/10.18653/v1/2020.wmt-1.1} {Findings of the 2020 conference on machine translation ({WMT}20)}.
\newblock In \emph{Proceedings of the Fifth Conference on Machine Translation}, pages 1--55, Online. Association for Computational Linguistics.

\bibitem[{Barrault et~al.(2019)Barrault, Bojar, Costa-juss{\`a}, Federmann, Fishel, Graham, Haddow, Huck, Koehn, Malmasi, Monz, M{\"u}ller, Pal, Post, and Zampieri}]{barrault-etal-2019-findings}
Lo{\"i}c Barrault, Ond{\v{r}}ej Bojar, Marta~R. Costa-juss{\`a}, Christian Federmann, Mark Fishel, Yvette Graham, Barry Haddow, Matthias Huck, Philipp Koehn, Shervin Malmasi, Christof Monz, Mathias M{\"u}ller, Santanu Pal, Matt Post, and Marcos Zampieri. 2019.
\newblock \href {https://doi.org/10.18653/v1/W19-5301} {Findings of the 2019 conference on machine translation ({WMT}19)}.
\newblock In \emph{Proceedings of the Fourth Conference on Machine Translation (Volume 2: Shared Task Papers, Day 1)}, pages 1--61, Florence, Italy. Association for Computational Linguistics.

\bibitem[{Berger et~al.(2024)Berger, Exel, Huck, and Riezler}]{berger-etal-2024-post}
Nathaniel Berger, Miriam Exel, Matthias Huck, and Stefan Riezler. 2024.
\newblock \href {https://doi.org/10.18653/v1/2024.wmt-1.122} {Post-edits are preferences too}.
\newblock In \emph{Proceedings of the Ninth Conference on Machine Translation}, pages 1289--1300, Miami, Florida, USA. Association for Computational Linguistics.

\bibitem[{Bojar et~al.(2017)Bojar, Chatterjee, Federmann, Graham, Haddow, Huang, Huck, Koehn, Liu, Logacheva, Monz, Negri, Post, Rubino, Specia, and Turchi}]{bojar-etal-2017-findings}
Ond{\v{r}}ej Bojar, Rajen Chatterjee, Christian Federmann, Yvette Graham, Barry Haddow, Shujian Huang, Matthias Huck, Philipp Koehn, Qun Liu, Varvara Logacheva, Christof Monz, Matteo Negri, Matt Post, Raphael Rubino, Lucia Specia, and Marco Turchi. 2017.
\newblock \href {https://doi.org/10.18653/v1/W17-4717} {Findings of the 2017 conference on machine translation ({WMT}17)}.
\newblock In \emph{Proceedings of the Second Conference on Machine Translation}, pages 169--214, Copenhagen, Denmark. Association for Computational Linguistics.

\bibitem[{Bojar et~al.(2018)Bojar, Federmann, Fishel, Graham, Haddow, Huck, Koehn, and Monz}]{bojar-etal-2018-findings}
Ond{\v{r}}ej Bojar, Christian Federmann, Mark Fishel, Yvette Graham, Barry Haddow, Matthias Huck, Philipp Koehn, and Christof Monz. 2018.
\newblock \href {https://doi.org/10.18653/v1/W18-6401} {Findings of the 2018 conference on machine translation ({WMT}18)}.
\newblock In \emph{Proceedings of the Third Conference on Machine Translation: Shared Task Papers}, pages 272--303, Belgium, Brussels. Association for Computational Linguistics.

\bibitem[{Buschbeck and Exel(2020)}]{buschbeck-exel-2020-parallel}
Bianka Buschbeck and Miriam Exel. 2020.
\newblock \href {https://doi.org/10.18653/v1/2020.wat-1.20} {A parallel evaluation data set of software documentation with document structure annotation}.
\newblock In \emph{Proceedings of the 7th Workshop on Asian Translation}, pages 160--169, Suzhou, China. Association for Computational Linguistics.

\bibitem[{Chi et~al.(2021)Chi, Dong, Wei, Yang, Singhal, Wang, Song, Mao, Huang, and Zhou}]{chi-etal-2021-infoxlm}
Zewen Chi, Li~Dong, Furu Wei, Nan Yang, Saksham Singhal, Wenhui Wang, Xia Song, Xian-Ling Mao, Heyan Huang, and Ming Zhou. 2021.
\newblock \href {https://doi.org/10.18653/v1/2021.naacl-main.280} {{I}nfo{XLM}: An information-theoretic framework for cross-lingual language model pre-training}.
\newblock In \emph{Proceedings of the 2021 Conference of the North American Chapter of the Association for Computational Linguistics: Human Language Technologies}, pages 3576--3588, Online. Association for Computational Linguistics.

\bibitem[{Christiano et~al.(2017)Christiano, Leike, Brown, Martic, Legg, and Amodei}]{christiano-etal-2017-deep}
Paul~F Christiano, Jan Leike, Tom Brown, Miljan Martic, Shane Legg, and Dario Amodei. 2017.
\newblock \href {https://proceedings.neurips.cc/paper_files/paper/2017/file/d5e2c0adad503c91f91df240d0cd4e49-Paper.pdf} {Deep reinforcement learning from human preferences}.
\newblock In \emph{Advances in Neural Information Processing Systems}, volume~30. Curran Associates, Inc.

\bibitem[{Conneau et~al.(2020)Conneau, Khandelwal, Goyal, Chaudhary, Wenzek, Guzm{\'a}n, Grave, Ott, Zettlemoyer, and Stoyanov}]{conneau-etal-2020-unsupervised}
Alexis Conneau, Kartikay Khandelwal, Naman Goyal, Vishrav Chaudhary, Guillaume Wenzek, Francisco Guzm{\'a}n, Edouard Grave, Myle Ott, Luke Zettlemoyer, and Veselin Stoyanov. 2020.
\newblock \href {https://doi.org/10.18653/v1/2020.acl-main.747} {Unsupervised cross-lingual representation learning at scale}.
\newblock In \emph{Proceedings of the 58th Annual Meeting of the Association for Computational Linguistics}, pages 8440--8451, Online. Association for Computational Linguistics.

\bibitem[{DiIanni and Deutsch(2025)}]{diianni-deutsch-2025-dont}
Colten DiIanni and Daniel Deutsch. 2025.
\newblock \href {https://doi.org/10.18653/v1/2025.emnlp-main.1273} {Don{'}t sweat the small stuff: Segment-level meta-evaluation based on pairwise difference correlation}.
\newblock In \emph{Proceedings of the 2025 Conference on Empirical Methods in Natural Language Processing}, pages 25062--25070, Suzhou, China. Association for Computational Linguistics.

\bibitem[{Freitag et~al.(2021)Freitag, Foster, Grangier, Ratnakar, Tan, and Macherey}]{freitag-etal-2021-mqm}
Markus Freitag, George Foster, David Grangier, Viresh Ratnakar, Qijun Tan, and Wolfgang Macherey. 2021.
\newblock \href {https://doi.org/10.1162/tacl_a_00437} {Experts, errors, and context: A large-scale study of human evaluation for machine translation}.
\newblock \emph{Transactions of the Association for Computational Linguistics}, 9:1460--1474.

\bibitem[{Guerreiro et~al.(2024)Guerreiro, Rei, van Stigt, Coheur, Colombo, and Martins}]{guerreiro-etal-2024-xcomet}
Nuno~M. Guerreiro, Ricardo Rei, Daan van Stigt, Luisa Coheur, Pierre Colombo, and Andr{\'e} F.~T. Martins. 2024.
\newblock \href {https://doi.org/10.1162/tacl_a_00683} {x{COMET}: Transparent machine translation evaluation through fine-grained error detection}.
\newblock \emph{Transactions of the Association for Computational Linguistics}, 12:979--995.

\bibitem[{Herbrich et~al.(2000)Herbrich, Graepel, and Obermayer}]{herbrich-etal-2000-large}
Ralf Herbrich, Thore Graepel, and Klaus Obermayer. 2000.
\newblock Large margin bank boundaries for ordinal regression.
\newblock In \emph{Advances in Large-Margin Classifiers}, pages 115--132. MIT Press.

\bibitem[{Hewitt(2021)}]{hewitt2021initializing}
John Hewitt. 2021.
\newblock \href {https:/nlp.stanford.edu/~johnhew//vocab-expansion.html} {Initializing new word embeddings for pretrained language models}.

\bibitem[{Kocmi et~al.(2024{\natexlab{a}})Kocmi, Avramidis, Bawden, Bojar, Dvorkovich, Federmann, Fishel, Freitag, Gowda, Grundkiewicz, Haddow, Karpinska, Koehn, Marie, Monz, Murray, Nagata, Popel, Popovi{\'c}, Shmatova, Steingr{\'i}msson, and Zouhar}]{kocmi-etal-2024-findings}
Tom Kocmi, Eleftherios Avramidis, Rachel Bawden, Ond{\v{r}}ej Bojar, Anton Dvorkovich, Christian Federmann, Mark Fishel, Markus Freitag, Thamme Gowda, Roman Grundkiewicz, Barry Haddow, Marzena Karpinska, Philipp Koehn, Benjamin Marie, Christof Monz, Kenton Murray, Masaaki Nagata, Martin Popel, Maja Popovi{\'c}, and 3 others. 2024{\natexlab{a}}.
\newblock \href {https://doi.org/10.18653/v1/2024.wmt-1.1} {Findings of the {WMT}24 general machine translation shared task: The {LLM} era is here but {MT} is not solved yet}.
\newblock In \emph{Proceedings of the Ninth Conference on Machine Translation}, pages 1--46, Miami, Florida, USA. Association for Computational Linguistics.

\bibitem[{Kocmi and Federmann(2023)}]{kocmi-federmann-2023-large}
Tom Kocmi and Christian Federmann. 2023.
\newblock \href {https://aclanthology.org/2023.eamt-1.19/} {Large language models are state-of-the-art evaluators of translation quality}.
\newblock In \emph{Proceedings of the 24th Annual Conference of the European Association for Machine Translation}, pages 193--203, Tampere, Finland. European Association for Machine Translation.

\bibitem[{Kocmi et~al.(2022)Kocmi, Matsushita, and Federmann}]{kocmi-etal-2022-ms}
Tom Kocmi, Hitokazu Matsushita, and Christian Federmann. 2022.
\newblock \href {https://doi.org/10.18653/v1/2022.wmt-1.47} {{MS}-{COMET}: More and better human judgements improve metric performance}.
\newblock In \emph{Proceedings of the Seventh Conference on Machine Translation (WMT)}, pages 541--548, Abu Dhabi, United Arab Emirates (Hybrid). Association for Computational Linguistics.

\bibitem[{Kocmi et~al.(2024{\natexlab{b}})Kocmi, Zouhar, Avramidis, Grundkiewicz, Karpinska, Popovi{\'c}, Sachan, and Shmatova}]{kocmi-etal-2024-error}
Tom Kocmi, Vil{\'e}m Zouhar, Eleftherios Avramidis, Roman Grundkiewicz, Marzena Karpinska, Maja Popovi{\'c}, Mrinmaya Sachan, and Mariya Shmatova. 2024{\natexlab{b}}.
\newblock \href {https://doi.org/10.18653/v1/2024.wmt-1.131} {Error span annotation: A balanced approach for human evaluation of machine translation}.
\newblock In \emph{Proceedings of the Ninth Conference on Machine Translation}, pages 1440--1453, Miami, Florida, USA. Association for Computational Linguistics.

\bibitem[{Kocmi et~al.(2024{\natexlab{c}})Kocmi, Zouhar, Federmann, and Post}]{kocmi-etal-2024-navigating}
Tom Kocmi, Vil{\'e}m Zouhar, Christian Federmann, and Matt Post. 2024{\natexlab{c}}.
\newblock \href {https://doi.org/10.18653/v1/2024.acl-long.110} {Navigating the metrics maze: Reconciling score magnitudes and accuracies}.
\newblock In \emph{Proceedings of the 62nd Annual Meeting of the Association for Computational Linguistics (Volume 1: Long Papers)}, pages 1999--2014, Bangkok, Thailand. Association for Computational Linguistics.

\bibitem[{Kudo and Richardson(2018)}]{kudo-richardson-2018-sentencepiece}
Taku Kudo and John Richardson. 2018.
\newblock \href {https://doi.org/10.18653/v1/D18-2012} {{S}entence{P}iece: A simple and language independent subword tokenizer and detokenizer for neural text processing}.
\newblock In \emph{Proceedings of the 2018 Conference on Empirical Methods in Natural Language Processing: System Demonstrations}, pages 66--71, Brussels, Belgium. Association for Computational Linguistics.

\bibitem[{Lin et~al.(2022)Lin, Kovacs, Shastry, Wuebker, and DeNero}]{lin-etal-2022-automatic}
Jessy Lin, Geza Kovacs, Aditya Shastry, Joern Wuebker, and John DeNero. 2022.
\newblock \href {https://doi.org/10.18653/v1/2022.naacl-main.36} {Automatic correction of human translations}.
\newblock In \emph{Proceedings of the 2022 Conference of the North American Chapter of the Association for Computational Linguistics: Human Language Technologies}, pages 494--507, Seattle, United States. Association for Computational Linguistics.

\bibitem[{Moghe et~al.(2025)Moghe, Fazla, Amrhein, Kocmi, Steedman, Birch, Sennrich, and Guillou}]{moghe-etal-2025-machine}
Nikita Moghe, Arnisa Fazla, Chantal Amrhein, Tom Kocmi, Mark Steedman, Alexandra Birch, Rico Sennrich, and Liane Guillou. 2025.
\newblock \href {https://doi.org/10.1162/coli_a_00537} {Machine translation meta evaluation through translation accuracy challenge sets}.
\newblock \emph{Computational Linguistics}, 51(1):73--137.

\bibitem[{{{Omnilingual MT Team}} et~al.(2025){{Omnilingual MT Team}}, Andrews, Artetxe, Meglioli, Costa-jussà, Chuang, Dale, Gao, Maillard, Mourachko, Ropers, Saleem, Sánchez, Tsiamas, Turkatenko, Ventayol-Boada, and Yates}]{teamBOUQuETDatasetBenchmark2025}
{{Omnilingual MT Team}}, Pierre Andrews, Mikel Artetxe, Mariano~Coria Meglioli, Marta~R. Costa-jussà, Joe Chuang, David Dale, Cynthia Gao, Jean Maillard, Alex Mourachko, Christophe Ropers, Safiyyah Saleem, Eduardo Sánchez, Ioannis Tsiamas, Arina Turkatenko, Albert Ventayol-Boada, and Shireen Yates. 2025.
\newblock \href {https://doi.org/10.48550/arXiv.2502.04314} {{{BOUQuET}}: Dataset, {{Benchmark}} and {{Open}} initiative for {{Universal Quality Evaluation}} in {{Translation}}}.
\newblock \emph{Preprint}, arXiv:2502.04314.

\bibitem[{Ouyang et~al.(2022)Ouyang, Wu, Jiang, Almeida, Wainwright, Mishkin, Zhang, Agarwal, Slama, Ray, Schulman, Hilton, Kelton, Miller, Simens, Askell, Welinder, Christiano, Leike, and Lowe}]{ouyang2022traininglanguagemodelsfollow}
Long Ouyang, Jeff Wu, Xu~Jiang, Diogo Almeida, Carroll~L. Wainwright, Pamela Mishkin, Chong Zhang, Sandhini Agarwal, Katarina Slama, Alex Ray, John Schulman, Jacob Hilton, Fraser Kelton, Luke Miller, Maddie Simens, Amanda Askell, Peter Welinder, Paul Christiano, Jan Leike, and Ryan Lowe. 2022.
\newblock \href {https://arxiv.org/abs/2203.02155} {Training language models to follow instructions with human feedback}.
\newblock \emph{Preprint}, arXiv:2203.02155.

\bibitem[{Popovi{\'c}(2021)}]{popovic-2021-agree}
Maja Popovi{\'c}. 2021.
\newblock \href {https://doi.org/10.18653/v1/2021.conll-1.18} {Agree to disagree: Analysis of inter-annotator disagreements in human evaluation of machine translation output}.
\newblock In \emph{Proceedings of the 25th Conference on Computational Natural Language Learning}, pages 234--243, Online. Association for Computational Linguistics.

\bibitem[{Proietti et~al.(2026)Proietti, Grundkiewicz, and Post}]{proietti-etal-2026-pear}
Lorenzo Proietti, Roman Grundkiewicz, and Matt Post. 2026.
\newblock \href {https://doi.org/10.18653/v1/2026.acl-long.1953} {{PEAR}: Pairwise evaluation for automatic relative scoring in machine translation}.
\newblock In \emph{Proceedings of the 64th Annual Meeting of the {A}ssociation for {C}omputational {L}inguistics (Volume 1: Long Papers)}, pages 42189--42207, San Diego, California, United States. Association for Computational Linguistics.

\bibitem[{Rafailov et~al.(2023)Rafailov, Sharma, Mitchell, Manning, Ermon, and Finn}]{rafailov-etal-2023-direct}
Rafael Rafailov, Archit Sharma, Eric Mitchell, Christopher~D Manning, Stefano Ermon, and Chelsea Finn. 2023.
\newblock \href {https://proceedings.neurips.cc/paper_files/paper/2023/file/a85b405ed65c6477a4fe8302b5e06ce7-Paper-Conference.pdf} {Direct preference optimization: Your language model is secretly a reward model}.
\newblock In \emph{Advances in Neural Information Processing Systems}, volume~36, pages 53728--53741. Curran Associates, Inc.

\bibitem[{Rei et~al.(2021)Rei, Farinha, Zerva, van Stigt, Stewart, Ramos, Glushkova, Martins, and Lavie}]{rei-etal-2021-references}
Ricardo Rei, Ana~C Farinha, Chrysoula Zerva, Daan van Stigt, Craig Stewart, Pedro Ramos, Taisiya Glushkova, Andr{\'e} F.~T. Martins, and Alon Lavie. 2021.
\newblock \href {https://aclanthology.org/2021.wmt-1.111/} {Are references really needed? unbabel-{IST} 2021 submission for the metrics shared task}.
\newblock In \emph{Proceedings of the Sixth Conference on Machine Translation}, pages 1030--1040, Online. Association for Computational Linguistics.

\bibitem[{Rei et~al.(2020)Rei, Stewart, Farinha, and Lavie}]{rei-etal-2020-comet}
Ricardo Rei, Craig Stewart, Ana~C Farinha, and Alon Lavie. 2020.
\newblock \href {https://doi.org/10.18653/v1/2020.emnlp-main.213} {{COMET}: A neural framework for {MT} evaluation}.
\newblock In \emph{Proceedings of the 2020 Conference on Empirical Methods in Natural Language Processing (EMNLP)}, pages 2685--2702, Online. Association for Computational Linguistics.

\bibitem[{Rei et~al.(2022)Rei, Treviso, Guerreiro, Zerva, Farinha, Maroti, C.~de Souza, Glushkova, Alves, Coheur, Lavie, and Martins}]{rei-etal-2022-cometkiwi}
Ricardo Rei, Marcos Treviso, Nuno~M. Guerreiro, Chrysoula Zerva, Ana~C Farinha, Christine Maroti, Jos{\'e}~G. C.~de Souza, Taisiya Glushkova, Duarte Alves, Luisa Coheur, Alon Lavie, and Andr{\'e} F.~T. Martins. 2022.
\newblock \href {https://doi.org/10.18653/v1/2022.wmt-1.60} {{C}omet{K}iwi: {IST}-unbabel 2022 submission for the quality estimation shared task}.
\newblock In \emph{Proceedings of the Seventh Conference on Machine Translation (WMT)}, pages 634--645, Abu Dhabi, United Arab Emirates (Hybrid). Association for Computational Linguistics.

\bibitem[{Ribeiro et~al.(2020)Ribeiro, Wu, Guestrin, and Singh}]{ribeiro-etal-2020-beyond}
Marco~Tulio Ribeiro, Tongshuang Wu, Carlos Guestrin, and Sameer Singh. 2020.
\newblock \href {https://doi.org/10.18653/v1/2020.acl-main.442} {Beyond accuracy: Behavioral testing of {NLP} models with {C}heck{L}ist}.
\newblock In \emph{Proceedings of the 58th Annual Meeting of the Association for Computational Linguistics}, pages 4902--4912, Online. Association for Computational Linguistics.

\bibitem[{Schmidt et~al.(2024)Schmidt, Reddy, Zhang, Alameddine, Uzan, Pinter, and Tanner}]{schmidt-etal-2024-tokenization}
Craig~W Schmidt, Varshini Reddy, Haoran Zhang, Alec Alameddine, Omri Uzan, Yuval Pinter, and Chris Tanner. 2024.
\newblock \href {https://doi.org/10.18653/v1/2024.emnlp-main.40} {Tokenization is more than compression}.
\newblock In \emph{Proceedings of the 2024 Conference on Empirical Methods in Natural Language Processing}, pages 678--702, Miami, Florida, USA. Association for Computational Linguistics.

\bibitem[{Schmidt et~al.(2026)Schmidt, Wahle, Ruas, and Gipp}]{schmidt-etal-2026-watches}
Finn Schmidt, Jan~Philip Wahle, Terry Ruas, and Bela Gipp. 2026.
\newblock \href {https://doi.org/10.18653/v1/2026.findings-acl.1145} {Who watches the watchmen? humans disagree with translation metrics on unseen domains}.
\newblock In \emph{Findings of the {A}ssociation for {C}omputational {L}inguistics: {ACL} 2026}, pages 22822--22841, San Diego, California, United States. Association for Computational Linguistics.

\bibitem[{Sharami et~al.(2023)Sharami, Shterionov, Blain, Vanmassenhove, Sisto, Emmery, and Spronck}]{sharami-etal-2023-tailoring}
Javad Pourmostafa~Roshan Sharami, Dimitar Shterionov, Fr{\'e}d{\'e}ric Blain, Eva Vanmassenhove, Mirella~De Sisto, Chris Emmery, and Pieter Spronck. 2023.
\newblock \href {https://aclanthology.org/2023.eamt-1.2/} {Tailoring domain adaptation for machine translation quality estimation}.
\newblock In \emph{Proceedings of the 24th Annual Conference of the European Association for Machine Translation}, pages 9--20, Tampere, Finland. European Association for Machine Translation.

\bibitem[{Snover et~al.(2006)Snover, Dorr, Schwartz, Micciulla, and Makhoul}]{snover-etal-2006-study}
Matthew Snover, Bonnie Dorr, Rich Schwartz, Linnea Micciulla, and John Makhoul. 2006.
\newblock \href {https://aclanthology.org/2006.amta-papers.25/} {A study of translation edit rate with targeted human annotation}.
\newblock In \emph{Proceedings of the 7th Conference of the Association for Machine Translation in the Americas: Technical Papers}, pages 223--231, Cambridge, Massachusetts, USA. Association for Machine Translation in the Americas.

\bibitem[{Tan and Monz(2025)}]{tan-monz-2025-remedy}
Shaomu Tan and Christof Monz. 2025.
\newblock \href {https://doi.org/10.18653/v1/2025.emnlp-main.217} {{R}e{M}edy: Learning machine translation evaluation from human preferences with reward modeling}.
\newblock In \emph{Proceedings of the 2025 Conference on Empirical Methods in Natural Language Processing}, pages 4370--4387, Suzhou, China. Association for Computational Linguistics.

\bibitem[{Uhlig et~al.(2025)Uhlig, Wuebker, Reinauer, and Denero}]{uhlig-etal-2025-cross}
Kaden Uhlig, Joern Wuebker, Raphael Reinauer, and John Denero. 2025.
\newblock \href {https://doi.org/10.18653/v1/2025.wmt-1.2} {Cross-lingual human-preference alignment for neural machine translation with direct quality optimization}.
\newblock In \emph{Proceedings of the Tenth Conference on Machine Translation}, pages 31--51, Suzhou, China. Association for Computational Linguistics.

\bibitem[{Wang et~al.(2022)Wang, Wu, and Neubig}]{wang-etal-2022-english}
Yaushian Wang, Ashley Wu, and Graham Neubig. 2022.
\newblock \href {https://doi.org/10.18653/v1/2022.emnlp-main.621} {{E}nglish contrastive learning can learn universal cross-lingual sentence embeddings}.
\newblock In \emph{Proceedings of the 2022 Conference on Empirical Methods in Natural Language Processing}, pages 9122--9133, Abu Dhabi, United Arab Emirates. Association for Computational Linguistics.

\bibitem[{Zouhar et~al.(2023)Zouhar, Dhuliawala, Zhou, Daheim, Kocmi, Jiang, and Sachan}]{zouhar-etal-2023-poor}
Vil{\'e}m Zouhar, Shehzaad Dhuliawala, Wangchunshu Zhou, Nico Daheim, Tom Kocmi, Yuchen~Eleanor Jiang, and Mrinmaya Sachan. 2023.
\newblock \href {https://doi.org/10.18653/v1/2023.eacl-main.95} {Poor man{'}s quality estimation: Predicting reference-based {MT} metrics without the reference}.
\newblock In \emph{Proceedings of the 17th Conference of the European Chapter of the Association for Computational Linguistics}, pages 1311--1325, Dubrovnik, Croatia. Association for Computational Linguistics.

\bibitem[{Zouhar et~al.(2024)Zouhar, Ding, Currey, Badeka, Wang, and Thompson}]{zouhar-etal-2024-fine}
Vil{\'e}m Zouhar, Shuoyang Ding, Anna Currey, Tatyana Badeka, Jenyuan Wang, and Brian Thompson. 2024.
\newblock \href {https://doi.org/10.18653/v1/2024.acl-short.45} {Fine-tuned machine translation metrics struggle in unseen domains}.
\newblock In \emph{Proceedings of the 62nd Annual Meeting of the Association for Computational Linguistics (Volume 2: Short Papers)}, pages 488--500, Bangkok, Thailand. Association for Computational Linguistics.

\end{thebibliography}

\appendix

\newpage

\section{Data Processing Details}\label{app:sec-data-proc}

We filter our internal data substantially before processing it.
For the initial filtering, we use these heuristics:

\begin{itemize}
    \item Only use documents with any human reviews
    \item Only use language pairs with $> 5,000$ segments available after document filtering
    \item Only using segments with $> 5$ but $< 425$ characters (around the 99$^{th}$ percentile in length)
\end{itemize}

We then process data into preference pairs on one hand and continuous scores on the other hand, as described below.
After applying both transformations separately, we sample train, dev, and test splits from both types of data, ensuring that each individual client project is assigned to only one split across both datasets to prevent overlap.

\subsection{Preference Pairs from Post-Edits}\label{subsec:internal-pref-pairs}

To create the preference pairs from our post-edit data, we filter for segments where the final translation was provided or approved by a human reviewer, treating this as the preferred translation.
As negative samples, we use previous machine-translated versions of the segment, where they differ from the final version.
Table~\ref{tab:dataset-sizes-internal} lists absolute test set sizes, and relative train and development set sizes across all language pairs used.
For the ACED corpus, we always use the post-edited segment as the preferred, provided it is different from the machine translated segment.
Because significant portions of the ACED data were not edited, this unfortunately limits the number of available preference pairs from this corpus (cf. Table~\ref{tab:data-sizes}).

\subsection{Continuous Scores from Post-Edits}\label{subsec:internal-scores}

Using HTER (human-targeted translation edit rate), as part of the training signal for metrics is an established practice \citep[e.g.,][]{sharami-etal-2023-tailoring}.
That said, TER is somewhat unstable on a segment level, especially when segments are short---a common occurrence in our data.
Thus, we select chrF++ as our surface metric for measuring post-edit distance.
Additionally, we use heuristics to identify numbers, URLs, and ``DNTs'' (\textbf{D}o \textbf{N}ot \textbf{T}ranslate) which the reviewer had to edit.
Since these are semantically major errors which may result only in a small edit distance, we subtract an additional 20 chrF points for these cases (equivalent to 5 out of a maximum score of 25 in MQM; see \citealp{freitag-etal-2021-mqm}).
Finally, we calculate z-scores from these artificial scores, and rescale them to between [0,1].

\subsection{WMT Data Processing}\label{app:wmt-data-proc}

To use the WMT data in both our training losses, we need both continuous scores and preference pairs.
As continuous scores, we use z-scores derived from the DA or ESA annotations, rescaled to between [0,1].
To transform the data into preference pairs, we compare the scores of all submissions for a given segment.
If the z-scores of two candidates are more than one standard deviation apart, we use them as a preference pair.
The raw scores act as a sanity check: We never use a submission as the rejected translation if its raw score is 98 or higher.
We subsample the WMT24 preference accuracy test set to one preference pair per source, in order to keep the test set size manageable.

\subsection{Quality Validation of Internal Data}\label{subsec:quality-val}

\begin{table}[t]
    \centering
    \begin{tabular}{lrrr}
    \toprule
    System & MT & AI-PE & Human \\
    \midrule
    MT & - & 0.26 & 0.14 \\
    AI-PE & 0.43 & - & 0.23 \\
    Human & 0.61 & 0.52 & - \\
    \bottomrule
    \end{tabular}
    \caption{Pairwise system comparison results from an internal study, where each cell indicates the win rate of the system listed in the row when compared against the system listed in the corresponding column. MT refers to machine translation, AI-PE to automatic post-editing of MT output, and Human to final human translation.}
    \label{tab:win-rates}
\end{table}

\begin{table}[t]
    \centering
    \begin{tabular}{lrrr}
    \toprule
    Metric & $\rho$ & $\tau$\\
    \midrule
    \cometbase & 0.404 & 0.282\\
    \cometkiwibase & 0.371 & 0.253 \\
    BLEU & 0.166 & 0.115\\
    chrF++ & 0.227 & 0.158\\
    TER & -0.195 & -0.137\\
    \bottomrule
    \end{tabular}
    \caption{Segment-level Spearman R ($\rho$) and Kendall’s Tau ($\tau$) correlations between ESA scores and automatic evaluation metrics in the internal study. BLEU, chrF++, and TER are computed using their segment-level variants. Only MT and AI-PE annotations are included, with the final human translation serving as the reference for reference-based metrics.}
    \label{tab:automatic-metric-correlations}
\end{table}

To provide a clearer sense of our internal data, we annotate a sample ($n=984$ annotations across $317$ source segments in $53$ documents) of English-German hypotheses in an ESA setting \citep{kocmi-etal-2024-error}.
The annotators in this small validation study are German-speaking NLP practitioners, and the data is sampled independently from \esadataacr.
The data sample is drawn randomly from our internal data, but with the criterion that there are three hypotheses present: A machine translation, an AI post-edit, and a final human post-edit.
Wherever possible, we show document context.

Table~\ref{tab:win-rates} shows how often each type of hypothesis wins against the other hypotheses.
That is, the human translation won over the MT system 61\% of the time while the MT system was rated preferable over the human translation 14\% of the time.
The rest are ties.
These win rates show how often our annotators agreed with the the edits made by the AI-PE or the human post-edit, effectively giving us a sense of how often edits are subjective.

Table~\ref{tab:automatic-metric-correlations} reports segment-level Spearman’s rank correlation ($\rho$) and Kendall’s Tau ($\tau$) between human judgements and automatic evaluation metrics.
Annotations corresponding to the final human post-edit are excluded and instead used as a reference, resulting in a total of $644$ annotations.
Overall, we observe slightly stronger correlations than those reported by \cite{rei-etal-2022-cometkiwi}.
While neural metrics show very similar correlation values, surface-level metrics show substantially higher correlations compared to prior work.
This can be attributed to the bias introduced by the post-editing setting in which the data was generated. Later edits tend to be close to the original version, which is measured by BLEU, TER and chrF++.

\section{Details of Tokeniser Modifications}\label{app:tok-details}

The pretrained encoder used by \cometkiwibase\
is based on XLM-R \citep{conneau-etal-2020-unsupervised}, specifically, InfoXLM \citep{chi-etal-2021-infoxlm}.
Thus, it uses the XLM-R tokeniser,
with a vocabulary of ca. 250k subword tokens, trained with SentencePiece \citep{kudo-richardson-2018-sentencepiece} and implemented for inference in the Huggingface \texttt{tokenizers} library.

By default, the XLM-R tokeniser splits on any whitespace, stripping away trailing and multiple spaces.
It also replaces spaces with the ``metaspace'' character.
Since we are only inferencing the tokeniser, it has already learned exclusively tokens that do not cross whitespace boundaries.
Rather than implement a new whitespace pretokeniser, we can simply remove this processing step to preserve trailing and multiple spaces without significantly changing the subword token sequence, as long as we keep the metaspace replacement.

Regarding the special space and control characters, the \texttt{tokenizers} library allows us to add tokens which bypass the Unicode normalisation step.
We add the non-breaking space (NBSP) in the narrow and regular form, the zero-width non-joiner (ZWNJ), the zero-width space (ZWSP) and control characters such as the left-to-right (LTR) and right-to-left (RTL) marker as additional tokens.
In the model's embedding layer, the new tokens are initialised according to the AvgEmb procedure shown by \citet{hewitt2021initializing}.

\section{MRL Regulariser and Margin Ablation}
\label{app:hyperparams}

\begin{figure*}
    \centering
    \includegraphics[width=0.9\linewidth]{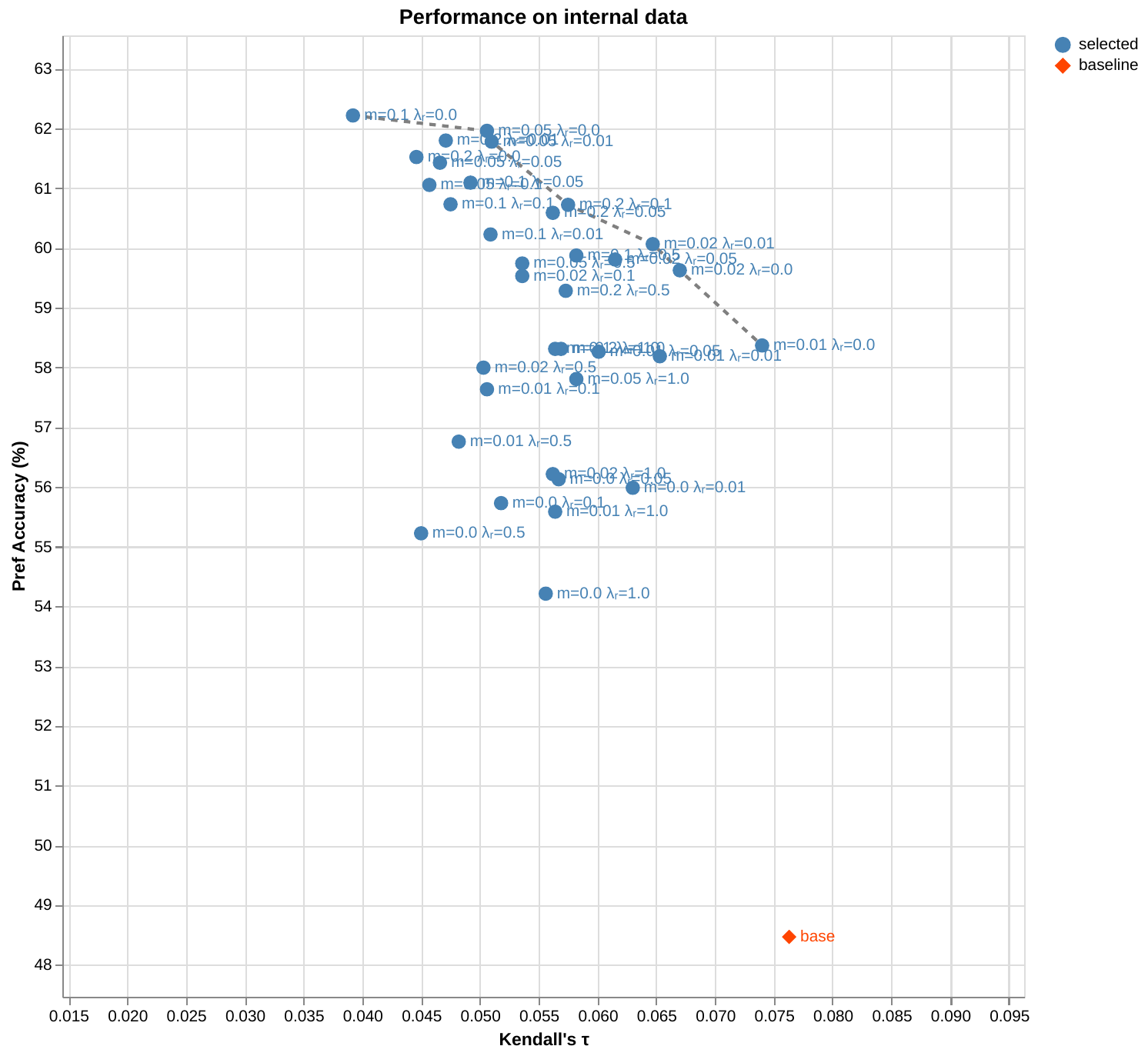}
    \caption{Pareto curve of Kendall's $\tau$ vs. preference accuracy achieved by contrastive training with different settings for $m$ and $\lambda$.}
    \label{fig:hyperparams}
\end{figure*}

Margin Ranking Loss comes with the \textit{margin} hyperparameter $m$.
Rather than set it only empirically, we test a range of margins in this section.

Additionally, we test a simple regularisation technique:
MRL on its own cannot provide a signal about appropriate absolute values.
However, computing a continuous score for both the positive and negative sample would require a held-out reference, which we do not have access to for our internal data.
Therefore, we compute the original model's score on both the positive and negative sample, and use these scores as a regularisation target.

Using mean-squared error (MSE) between the predicted score $x_i$ and the original score $y_i$, as a regulariser, the full loss equation becomes:

\begin{equation}
\begin{aligned}
    \mathcal{L} & = \text{MRL}(x_{pos},x_{neg}) \\
    & + \lambda_r (\text{MSE}(x_{pos}, y_{pos}) + \text{MSE}(x_{neg},y_{neg}))
\end{aligned}
\end{equation}

We ablate the hyperparameters $m$ and $\lambda$, testing margins $m \in \{0.0, 0.01, 0.02, 0.05, 0.1 \}$, and regulariser weights $\lambda \in \{0.0, 0.01, 0.05, 0.1, 0.5, 1\}$.
Figure~\ref{fig:hyperparams} shows the resulting pareto curve over two metrics: Segment-wise preference accuracy and global Kendall's $\tau$, both measured on our internal test set.

Since \cometkiwibase\ starts out with poor preference accuracy on our internal data, this regulariser frequently conflicts with MRL.
Tuning the strength of the regularisation loss via $\lambda$ proves insufficient for balancing the two: The majority of points on the pareto curve has a $\lambda = 0.0$.
Therefore, we discard regularisation to the original scores, and instead focus on multi-task training.

We select a margin of 0.02, which sits on the pareto curve, and corresponds to the common semantics of COMET-type metrics, where a difference of 0.02 already corresponds to a meaningful difference in quality \citep[cf.][]{kocmi-etal-2024-navigating}.

\section{Data Mix Ablation}
\label{app:data-mix-ablation}

In the multi-task setup, we balance the influence of the two losses by varying the composition of the training data. 
This leaves open two choices: the share of original training data (WMT2017-20) mixed into the training data, and the ratio between continuous scores (MSE) and preference pairs (MRL).
Table~\ref{tab:data-mix-ablation} shows our ablation experiment for both choices.

First, we vary the share of internal data from 0\% to 100\%, applying the same mix to the MSE and the MRL side (10k samples each).
We originally mixed in WMT data to mitigate forgetting (cf.~\S~\ref{sec:datasets}).
In the multi-task setting, however, preference accuracy improves with a growing share of internal data across all preference-based test sets, including WMT24 itself.
Only ACES and the global correlations benefit from added WMT data, mirroring the trade-off between segment-wise and global performance discussed in the main text.
We decide to train our final models on internal data only.

Then, we vary the ratio of MSE to MRL samples on this internal-only mix, from pure MRL to pure MSE at a matched total budget of 15k samples.
The two extremes recover the single-loss failure modes: Pure MRL yields the highest preference accuracy on the internal test set, but collapses the global correlations and ACES performance, while pure MSE shows the opposite behaviour.
Between the extremes, the measures trade off smoothly.
We select the 2:1 ratio, which achieves the best WMT24 accuracy and strong \locchecklist\ performance, and retains most of the global-correlation gains of MSE-only training, at a small cost in preference accuracy on the internal test set.

\begin{table*}[htb]
    \centering
    \small
    \resizebox{\textwidth}{!}{
    \begin{tabular}{lrrrrrrr}
    \toprule
        & \multicolumn{2}{c}{\textbf{\locchecklist}} & \multicolumn{2}{c}{\textbf{Internal Data}} & \textbf{\esadataacr} & \textbf{WMT24} & \textbf{ACES} \\
        \textbf{Model} & \textbf{Internal} & \textbf{Public} & \textbf{Pref Acc.} & \textbf{Pearson} & \textbf{Kendall} & \textbf{Pref Acc.} & \textbf{Score} \\
    \midrule
        Baseline & 45.1 & 30.1 & 53.1 & 0.173 & 0.107 & 66.2 & 17.94 \\
    \midrule
        \multicolumn{8}{l}{\emph{Internal/WMT data mix} (MSE 10k + MRL 10k)} \\
    \midrule
        WMT only & 52.7 & 36.7 & 52.7 & 0.201 & 0.120 & 66.2 & \textbf{17.40} \\
        25\% internal & 58.9 & — & 56.2 & 0.222 & \textbf{0.138} & 66.9 & 14.16 \\
        50\% internal & 61.0 & 40.8 & 55.2 & 0.220 & 0.135 & 66.9 & 15.08 \\
        75\% internal & 61.9 & — & 58.4 & \textbf{0.225} & 0.124 & 66.8 & 13.14 \\
        Internal only & \textbf{62.1} & \textbf{44.4} & \textbf{60.7} & 0.194 & 0.131 & \textbf{67.5} & 11.25 \\
    \midrule
        \multicolumn{8}{l}{\emph{MSE:MRL supervision ratio} (internal data only)} \\
    \midrule
        0:1 (MRL only, 15k) & 54.0 & \textbf{54.2} & \textbf{66.9} & 0.113 & 0.044 & 63.9 & 3.67 \\
        1:4 (2.5k + 10k) & 61.0 & 44.5 & 61.2 & 0.179 & 0.110 & 67.0 & 11.20 \\
        1:2 (5k + 10k) & 61.1 & 46.7 & 61.4 & 0.178 & 0.112 & 67.5 & 11.50 \\
        1:1 (10k + 10k) & 62.1 & 44.4 & 60.7 & 0.194 & 0.131 & 67.5 & 11.25 \\
        2:1 (10k + 5k, ours) & 64.1 & 46.3 & 59.6 & 0.207 & 0.135 & \textbf{67.6} & 11.97 \\
        4:1 (10k + 2.5k) & \textbf{65.4} & 45.9 & 56.5 & 0.218 & 0.127 & 67.1 & 13.25 \\
        1:0 (MSE only, 15k) & 61.6 & 41.4 & 51.1 & \textbf{0.234} & \textbf{0.145} & 65.5 & \textbf{14.58} \\
    \bottomrule
    \end{tabular}
    }
    \caption{Ablation of training data mix. We use varying ratios of a fixed pool of data. We conclude that the multi-task training sees limited benefit from replaying WMT data, and settle on a 2:1 ratio of the two losses.}
    \label{tab:data-mix-ablation}
\end{table*}

\section{\locchecklist\ Details}
\label{app:loc-checklist}

\begin{table*}[htb]
    \centering
    \small
     \resizebox{\textwidth}{!}{
    \begin{tabular}{lrrrrrrrrrrrr}
    \toprule
         & \textbf{en-ar} & \textbf{en-de} & \textbf{en-es} & \textbf{en-fi} & \textbf{en-fr} & \textbf{en-hi} & \textbf{en-ja} & \textbf{en-pl} & \textbf{en-ru} & \textbf{en-th} & \textbf{en-zh-Hans} & \textbf{all} \\
         \midrule
         Copy URLs & -- & 8 & 6 & -- & 12 & -- & 18 & 2 & 1 & -- & 5 & 52 \\
         Copy DNTs & -- & -- & -- & -- & -- & -- & -- & 1 & 4 & -- & -- & 5 \\
         Long Numbers & 50 & 50 & 50 & 50 & 50 & 50 & 50 & 50 & 50 & 34 & 50 & 534 \\
         Lead./Trail. Sp. & 50 & 50 & 50 & 50 & 50 & 50 & 50 & 50 & 50 & 50 & 50 & 550 \\
         Extra Period & 50 & 50 & 50 & 50 & 50 & 50 & 50 & 50 & 50 & -- & 50 & 500 \\
         All Caps & -- & 50 & 50 & 50 & 50 & -- & -- & 50 & 50 & -- & -- & 300 \\
         Spanish ¿ \& ¡ & -- & -- & 50 & -- & -- & -- & -- & -- & -- & -- & -- & 50 \\
         French NBSPs & -- & -- & -- & -- & 50 & -- & -- & -- & -- & -- & -- & 50 \\
         Unicode Spaces & 50 & 50 & 50 & 50 & 50 & 39 & 50 & 50 & 50 & 34 & 50 & 523 \\
         \midrule
         \textbf{Totals} & 200 & 258 & 306 & 250 & 312 & 189 & 218 & 253 & 255 & 118 & 205 & \textbf{2,564} \\
    \bottomrule
    \end{tabular}
    }
    \caption{Distribution of samples in the internal \locchecklist\ split.}
    \label{tab:challenge-set-comp}
\end{table*}

\begin{table*}[htb]
    \centering
    \small
    \begin{tabular}{lrrrrrrrrr}
    \toprule
     & \textbf{en-ar} & \textbf{en-de} & \textbf{en-es} & \textbf{en-fr} & \textbf{en-hi} & \textbf{en-ja} & \textbf{en-ru} & \textbf{en-zh} & \textbf{all} \\
\midrule
         Copy URLs & --& -- & -- & -- & -- & -- & -- & -- & -- \\
         Copy DNTs & -- & -- & -- & -- & -- & -- & -- & -- & -- \\
        Long Numbers & 0 & 3 & 2 & 2 & 6 & 2 & 5 & 0 & 20 \\
        Lead./Trail. Sp. & 12 & 0 & 2 & 0 & 50 & 42 & 46 & 8 & 160 \\
        Extra Period & 50 & 30 & 31 & 30 & 50 & 50 & 50 & 50 & 341 \\
All Caps & 0 & 50 & 50 & 50 & 0 & 0 & 50 & 0 & 200 \\
Spanish ¿ \& ¡  & 0 & 0 & 50 & 0 & 0 & 0 & 0 & 0 & 50 \\
French NBSPs & 0 & 0 & 0 & 50 & 0 & 0 & 0 & 0 & 50 \\
Unicode Spaces & 0 & 0 & 0 & 18 & 0 & 2 & 0 & 0 & 20 \\
\midrule
\textbf{Totals} & 62 & 83 & 135 & 150 & 106 & 96 & 151 & 58 & \textbf{841} \\

    \bottomrule
    \end{tabular}
    \caption{Distribution of samples across categories in the public \locchecklist\ split.
    Samples are distributed across the eight \locchecklist\ language pairs that occur in WMT25 and BOUQuET.}
    \label{tab:challenge-set-comp-public}
\end{table*}

\subsection{Composition}\label{app:loc-checklist-comp}

We include eleven language pairs in \locchecklist, translating from English into the following target languages: Arabic, German, Spanish, Finnish, French, Hindi, Japanese, Polish, Russian, Thai, and Chinese.
These target languages cover both very high-resource and more mid-resource language pairs, seven different scripts, and as many language families.
We apply the following heuristics to find and create positive and negative samples:

\begin{table*}[htb]
    \centering
    \small
    \resizebox{\textwidth}{!}{%
    \begin{tabular}{lrrrrrrrr}
    \toprule
        & \textbf{\cometkiwibase} & \textbf{\texttt{+Tok}} & \textbf{\texttt{+MSE}} & \textbf{\texttt{+Tok+MSE}} & \textbf{\texttt{+MRL}} & \textbf{\texttt{+Tok+MRL}} & \textbf{\texttt{+Tok+Multi}} & \textbf{\texttt{+Tok+Multi+Aug}} \\
    \midrule
Copy URLs & 78.8 & 80.8 & \textbf{82.7} & 78.8 & 34.6 & 28.8 & 76.9 & 78.8 \\
Copy DNTs & \textbf{100.0} & \textbf{100.0} & \textbf{100.0} & \textbf{100.0} & \textbf{100.0} & \textbf{100.0} & \textbf{100.0} & \textbf{100.0} \\
Long Numbers & 85.2 & 81.3 & \textbf{97.9} & 96.8 & 37.5 & 33.1 & 89.3 & 97.0 \\
Lead./Trail. Sp. & 25.5 & 68.4 & 27.5 & 44.4 & 31.6 & 41.5 & 39.3 & \textbf{78.2} \\
Extra Period & 40.2 & 44.6 & 51.8 & 47.6 & 44.6 & 38.6 & 49.4 & \textbf{92.6} \\
All Caps & 22.3 & 26.7 & 64.3 & 63.3 & 63.7 & 76.7 & 68.3 & \textbf{99.7} \\
Spanish ¿ \& ¡ & \textbf{96.0} & \textbf{96.0} & 78.0 & 72.0 & \textbf{96.0} & 92.0 & 82.0 & 90.0 \\
French NBSPs & 42.0 & 8.0 & 50.0 & 38.0 & 52.0 & 82.0 & 80.0 & \textbf{88.0} \\
Unicode Spaces & 34.2 & 11.1 & 39.2 & 55.4 & 36.3 & 85.9 & 71.1 & \textbf{86.6} \\
    \bottomrule
    \end{tabular}%
    }
   \caption{Performance of COMETKiwi models on the Localisation CheckList set by error category. Fine-tuning was done with the internal dataset throughout.}
   \label{tab:challenge-set-perf}
\end{table*}

\begin{itemize}
    \item \textbf{Copy URLs.} Although URLs may be localised, an MT model cannot know the localised URL and so should be expected to copy the source URL if no other instruction is given.
    We sample translations where a URL was changed during the post-editing process.
    We ensure that the reference and positive sample contain the source URL, then place the changed URL in the negative sample.
    \item \textbf{Copy DNTs.} \textbf{D}o \textbf{N}ot \textbf{T}ranslate items are elements within the segment, usually marked by a specific syntax, that should be copied to the correct place in the target segment, and must not be translated.
    We sample segments where a DNT item was changed during the post-editing process, analogously to URLs.
    \item \textbf{Long Numbers.} If a number is mistranslated, the metric should catch this and consider it a major error, even if the number is long. We sample translations organically containing long digit sequences, and check that the intermediate translation matches the reference. Then, we randomly remove, add, or swap one or more digits, and place the resulting digit sequence within the negative sample.
    \item \textbf{Leading/Trailing Spaces.} Translators are careful to leave leading and trailing spaces in a segment unchanged, but the default tokenisation of COMETKiwi models drops these entirely. We sample translations with leading or trailing spaces, then randomly add, remove, or replace whitespace characters from these leading or trailing space sequences. The whitespace characters we sample from for addition or replacement are the regular space, tab, NBSP, ZWSP, and ZWNJ.
    \item \textbf{Extra Period.} If a longer segment does not end in a period, language models still sometimes prefer a translation ending in a period, but translators do not. We sample segments with a minimum of ten source words that do not end in a period, then add the appropriate sentence-ending symbol for each script to the negative sample.
    \item \textbf{All Caps.} If the source segment is written in all-caps, a translation in all-caps should be preferred to a translation in sentence case. We sample segments in sentence case and use the initial translation as the negative sample, then upper-case the source, positive sample, and reference. This criterion is only applicable in Latin and Cyrillic script.
    \item \textbf{Spanish ¿ and ¡.} We want to verify that the sentence-initial exclamation mark and question mark in Spanish are treated correctly by the metrics. We sample Spanish sentences containing these characters and remove them to create the negative sample.
    \item \textbf{French NBSPs.} French has specific rules about where non-breaking spaces should be used, but COMETKiwi models cannot differentiate special space characters. We sample French segments where a NBSP occurs with punctuation, and randomly either remove the NBSP or replace it with a regular whitespace character.
    \item \textbf{Unicode Spaces.} Some segments contain special Unicode whitespace and control characters such as NBSP, narrow NBSP, zero-width space, zero-width non-joiner (ZWNJ), zero-width joiner (ZWJ), ideographic space, and the LTR/RTL marks. These are stripped by the original tokenizer but often need to be kept in the translation. We find segments where the reference contains any of these characters. We use the initial translation as a positive example if it contains the characters as well; otherwise, we use the reference. We produce the negative example by either dropping the special character or replacing it with a regular space.
\end{itemize}

Table~\ref{tab:challenge-set-comp} shows the composition of the Localisation CheckList by language pair and error category.

\subsection{Analysis of Results}\label{subsec:challenge-set-results}

Table~\ref{tab:challenge-set-perf} breaks down the performance of COMETKiwi variants, fine-tuned on internal data, by category.
Long numbers show a strong improvement under MSE fine-tuning.
Recall that we explicitly subtracted from the chrF scores assigned to segments failing this criterion, so the QE model has clearly learned to attend to this pattern due to this data augmentation.

As expected, the category of leading and trailing spaces is strongly helped by allowing the model's tokeniser to see the whitespace in question.
Fine-tuning harms performance here, likely not providing a useful signal for this category.
After adding augmented examples to the training data the model learns to recognize the leading and trailing spaces difference with 78.2\% accuracy.

Regarding the NBSP, merely adding it to the tokeniser harms performance because there is now a new, untrained token in many of the segments.
Fine-tuning with MSE helps but does not restore performance to baseline---chrF++ does not distinguish the NBSP either.
Here, however, contrastive and multi-task fine-tuning proves effective: Our localisation data clearly contains a number of segments where this type of edit was made, and consistently so.
Adding augmented training data further raises performance to 88\%.
A similar picture emerges for the broader Unicode Spaces rule.

Accuracy on the `Extra Period' category stays in a similar range across training approaches, again suggesting a scarcity of useful signal.
This is supported by the result after adding augmentation, which helps the model to perfectly distinguish this phenomenon very well.
Additionally, fine-tuning is relatively effective at making the model rate an all-caps translation of an all-caps source above the sentence-case version, but data augmentation is once again necessary to consistently demonstrate the rule.

The Spanish punctuation marks stand out somewhat: The base QE model already shows very high performance on this category.
Fine-tuning with MSE hurts compared to the baseline, and even the multi-task training with data augmentation stays slightly behind the baseline, but this is likely noise due to a small sample size.

\section{Additional Results}
\label{app:extra-tables}

We check the general capabilities of our fine-tuned models on the ACES challenge set \citep{amrhein-etal-2023-aces}, primarily tracking the aggregated ACES-Score.
Table~\ref{tab:results-gen-cap} shows the ACES-Score \citep{amrhein-etal-2023-aces} achieved by fine-tuned metrics, a proxy for their general capabilities.
It also lists their preference accuracy on the WMT24 test set.
Table~\ref{tab:dataset-sizes-internal} lists the dataset sizes of our internal data per language pair, while Table~\ref{tab:results-pref_acc-segment-history-test-prefs} shows per-language-pair performance of fine-tuned metric and QE models on that same data.
Tables~\ref{tab:results-kendall_corr-snap_internal} and~\ref{tab:results-kendall_corr-snap_aced} break down the per-language-pair performance of our QE models on \esadataacr.

\begin{table}[ht]
\centering
\small
\begin{tabular}{lrr}
\toprule
\textbf{Model} & \textbf{ACES} & \textbf{WMT24} \\
\midrule
GEMBA-ESA (gpt-4.1) & 10.41 & 63.8 \\
\midrule
\cometkiwibase & \textbf{17.94} & 66.2 \\
+ New Tok & 17.88 & 66.2 \\
\midrule
\textit{Internal data fine-tuning} \\
\midrule
+ MSE & 13.37 & 65.9 \\
+ New Tok + MSE & 14.58 & 65.5 \\
+ MRL & 4.34 & 64.5 \\
+ New Tok + MRL & 3.67 & 63.9 \\
+ New Tok + MSE + MRL & 11.97 & 67.6 \\
+ New Tok + MSE + MRL + Aug & 12.60 & 67.2 \\
\midrule
\textit{ACED fine-tuning} \\
\midrule
+ MSE & 13.56 & 65.8 \\
+ New Tok + MSE & 14.37 & 66.2 \\
+ MRL & 17.24 & \textbf{67.8} \\
+ New Tok + MRL & 17.13 & 67.5 \\
+ New Tok + MSE + MRL & 13.52 & 66.2 \\
+ New Tok + MSE + MRL + Aug & 14.01 & 66.0 \\
\bottomrule

\end{tabular}
\caption{General metric capabilities, in terms of ACES-Score and preference accuracy (in percent) on WMT24.}
\label{tab:results-gen-cap}
\end{table}

\begin{table*}[ht]
\centering
\small
\begin{tabular}{lrrrrrrrrr}
\toprule
\textbf{Dataset} & \textbf{de-en} & \textbf{de-es} & \textbf{de-fr} & \textbf{en-ar} & \textbf{en-bg} & \textbf{en-bn} & \textbf{en-cs} & \textbf{en-da} & \textbf{en-de} \\
\midrule
Train (Continuous Scores) & 0.48\% & 0.20\% & 0.21\% & 3.58\% & 0.70\% & 0.27\% & 1.93\% & 2.76\% & 4.60\% \\
Train (Preferences) & 0.37\% & 0.06\% & 0.05\% & 3.53\% & 0.45\% & 0.24\% & 1.62\% & 1.85\% & 4.81\% \\
Dev (Continuous Scores) & 0.57\% & -- & -- & 1.55\% & 0.04\% & 0.13\% & 0.55\% & 1.70\% & 2.84\% \\
Dev (Preferences) & 0.13\% & -- & 0.32\% & 1.58\% & 0.58\% & 0.17\% & 5.32\% & 1.26\% & 1.45\% \\
Test (Continuous Scores) & 81 & 500 & 46 & 500 & -- & 500 & 500 & 500 & 500 \\
Test (Preferences) & 185 & 500 & 24 & 500 & -- & 500 & 500 & 500 & 500 \\
\bottomrule
\end{tabular}
\vspace{0.5em}
\begin{tabular}{lrrrrrrrrr}
\toprule
\textbf{Dataset} & \textbf{en-el} & \textbf{en-es} & \textbf{en-fa} & \textbf{en-fi} & \textbf{en-fr} & \textbf{en-he} & \textbf{en-hi} & \textbf{en-hr} & \textbf{en-hu} \\
\midrule
Train (Continuous Scores) & 1.10\% & 5.09\% & 0.32\% & 5.66\% & 4.57\% & 0.47\% & 1.77\% & 0.84\% & 0.57\% \\
Train (Preferences) & 0.68\% & 5.00\% & 0.21\% & 6.03\% & 5.15\% & 0.67\% & 1.91\% & 1.42\% & 1.15\% \\
Dev (Continuous Scores) & 2.04\% & 1.90\% & 0.50\% & 0.09\% & 2.51\% & 2.32\% & 1.47\% & 0.19\% & 1.89\% \\
Dev (Preferences) & 2.57\% & 4.88\% & 0.01\% & 0.32\% & 6.65\% & 1.02\% & 0.07\% & 0.14\% & 1.69\% \\
Test (Continuous Scores) & 500 & 500 & -- & 500 & 500 & 500 & 500 & -- & 258 \\
Test (Preferences) & 500 & 500 & -- & 500 & 500 & 500 & 500 & -- & 110 \\
\bottomrule
\end{tabular}
\vspace{0.5em}
\begin{tabular}{lrrrrrrrrr}
\toprule
\textbf{Dataset} & \textbf{en-hy} & \textbf{en-id} & \textbf{en-it} & \textbf{en-ja} & \textbf{en-ko} & \textbf{en-lt} & \textbf{en-lv} & \textbf{en-mr} & \textbf{en-ms} \\
\midrule
Train (Continuous Scores) & 0.10\% & 1.85\% & 4.04\% & 4.40\% & 3.97\% & 0.26\% & 0.25\% & 0.19\% & 1.29\% \\
Train (Preferences) & 0.30\% & 1.25\% & 5.43\% & 4.56\% & 4.06\% & 0.31\% & 0.01\% & 0.19\% & 0.53\% \\
Dev (Continuous Scores) & -- & 1.74\% & 9.11\% & 6.77\% & 7.57\% & 0.43\% & 0.01\% & 1.25\% & -- \\
Dev (Preferences) & -- & 6.61\% & 3.61\% & 7.69\% & 14.06\% & 0.02\% & 0.33\% & -- & 0.47\% \\
Test (Continuous Scores) & -- & 500 & 500 & 500 & 500 & 95 & 500 & 5 & 61 \\
Test (Preferences) & -- & 500 & 500 & 500 & 500 & 114 & 500 & 10 & 41 \\
\bottomrule
\end{tabular}
\vspace{0.5em}
\begin{tabular}{lrrrrrrrrr}
\toprule
\textbf{Dataset} & \textbf{en-nl} & \textbf{en-no} & \textbf{en-pl} & \textbf{en-pt} & \textbf{en-ro} & \textbf{en-ru} & \textbf{en-sk} & \textbf{en-sl} & \textbf{en-so} \\
\midrule
Train (Continuous Scores) & 4.78\% & 2.97\% & 4.38\% & 4.23\% & 2.18\% & 4.18\% & 0.46\% & 0.31\% & 0.25\% \\
Train (Preferences) & 4.44\% & 2.27\% & 3.52\% & 4.77\% & 2.20\% & 3.75\% & 0.50\% & 0.27\% & 0.39\% \\
Dev (Continuous Scores) & 5.29\% & 1.58\% & 4.12\% & 3.43\% & 5.64\% & 0.85\% & 1.34\% & -- & -- \\
Dev (Preferences) & 5.52\% & 5.73\% & 2.52\% & 3.09\% & 10.64\% & 0.83\% & 0.77\% & 0.30\% & -- \\
Test (Continuous Scores) & 500 & 500 & 500 & 500 & 500 & 500 & 500 & -- & -- \\
Test (Preferences) & 500 & 500 & 500 & 500 & 314 & 500 & 419 & -- & -- \\
\bottomrule
\end{tabular}
\vspace{0.5em}
\begin{tabular}{lrrrrrrrrr}
\toprule
\textbf{Dataset} & \textbf{en-sq} & \textbf{en-sv} & \textbf{en-sw} & \textbf{en-ta} & \textbf{en-te} & \textbf{en-th} & \textbf{en-tl} & \textbf{en-tr} & \textbf{en-uk} \\
\midrule
Train (Continuous Scores) & 0.15\% & 2.49\% & 0.01\% & 0.18\% & 0.25\% & 3.84\% & 0.86\% & 1.56\% & 3.74\% \\
Train (Preferences) & 0.25\% & 1.83\% & 0.37\% & 0.18\% & 0.27\% & 4.01\% & 0.73\% & 2.41\% & 3.05\% \\
Dev (Continuous Scores) & -- & 7.89\% & -- & 0.67\% & -- & 4.32\% & -- & 6.70\% & 0.55\% \\
Dev (Preferences) & 0.00\% & 1.14\% & -- & 0.12\% & -- & 1.67\% & 0.02\% & 0.30\% & 0.25\% \\
Test (Continuous Scores) & -- & 500 & 97 & -- & 373 & 500 & -- & 500 & 500 \\
Test (Preferences) & -- & 500 & 62 & -- & 367 & 500 & -- & 500 & 500 \\
\bottomrule
\end{tabular}
\vspace{0.5em}
\begin{tabular}{lrrrrrrrrr}
\toprule
\textbf{Dataset} & \textbf{en-ur} & \textbf{en-vi} & \textbf{en-zh-Hans} & \textbf{en-zh-Hant} & \textbf{es-en} & \textbf{pt-en} &  &  & \textbf{Total} \\
\midrule
Train (Continuous Scores) & 0.16\% & 2.07\% & 4.35\% & 3.71\% & 1.25\% & 0.14\% &  &  & 100.0\% \\
Train (Preferences) & 0.16\% & 1.49\% & 5.48\% & 4.18\% & 1.22\% & 0.43\% &  &  & 100.0\% \\
Dev (Continuous Scores) & 0.74\% & 0.34\% & 3.65\% & 2.69\% & 2.87\% & 0.14\% &  &  & 100.0\% \\
Dev (Preferences) & 0.59\% & 1.16\% & 1.55\% & 2.86\% & 0.00\% & -- &  &  & 100.0\% \\
Test (Continuous Scores) & -- & 500 & 500 & 500 & 500 & 381 &  &  & 17,397 \\
Test (Preferences) & -- & 500 & 500 & 500 & 327 & 160 &  &  & 16,633 \\
\bottomrule
\end{tabular}
\caption{Internal dataset sizes by language pair.}
\label{tab:dataset-sizes-internal}
\end{table*}

\begin{table*}[htb]
\centering
\small
\resizebox{\textwidth}{!}{%
\begin{tabular}{lrrrrrrrrrr}
\toprule
\textbf{Model} & \textbf{en-ar} & \textbf{en-de} & \textbf{en-es} & \textbf{en-fi} & \textbf{en-fr} & \textbf{en-ja} & \textbf{en-pl} & \textbf{en-pt} & \textbf{en-sv} & \textbf{en-tr} \\
\midrule
\cometkiwibase & 52.2 & 54.2 & 51.4 & 54.4 & 48.6 & 49.0 & 50.4 & 50.0 & 55.0 & 51.6 \\
+ New Tok & 52.8 & 53.4 & 52.4 & 53.6 & 47.2 & 50.8 & 45.4 & 49.4 & 54.0 & 51.0 \\
+  MSE & 54.0 & 49.0 & 43.8 & 48.0 & 42.8 & 53.0 & 50.4 & 46.2 & 49.2 & 47.6 \\
+ New Tok +  MSE & 51.8 & 47.2 & 44.0 & 48.8 & 43.0 & 51.2 & 51.0 & 44.0 & 48.4 & 48.0 \\
+  MRL & 73.2 & \textbf{73.6} & 65.0 & 64.6 & \textbf{70.0} & 66.6 & 67.2 & 72.4 & 63.4 & 57.6 \\
+ New Tok +  MRL & \textbf{73.8} & 73.2 & \textbf{67.8} & \textbf{65.2} & \textbf{70.0} & \textbf{68.6} & \textbf{71.6} & \textbf{74.6} & \textbf{65.4} & \textbf{59.6} \\
+ New Tok +  Multi & 61.6 & 58.0 & 55.0 & 55.8 & 54.8 & 63.4 & 59.8 & 59.0 & 51.4 & 56.4 \\
+ New Tok +  Multi + Aug & 59.4 & 53.2 & 51.6 & 51.2 & 52.2 & 61.8 & 54.2 & 55.2 & 49.6 & 52.6 \\
\bottomrule
\end{tabular}%
}
\caption{Finetuning on internal data---Preference accuracy on internal data, by language pair.}
\label{tab:results-pref_acc-segment-history-test-prefs}
\end{table*}

\begin{table*}[htb]
\centering
\small
\resizebox{\textwidth}{!}{%
\begin{tabular}{lrrrrrrrrrr}
\toprule
\textbf{Model} & \textbf{en-ar} & \textbf{en-de} & \textbf{en-es} & \textbf{en-fi} & \textbf{en-fr} & \textbf{en-ja} & \textbf{en-pl} & \textbf{en-pt} & \textbf{en-sv} & \textbf{en-tr} \\
\midrule
\cometkiwibase & 52.2 & 54.2 & 51.4 & 54.4 & 48.6 & 49.0 & 50.4 & 50.0 & 55.0 & \textbf{51.6} \\
+ New Tok & 52.8 & 53.4 & 52.4 & 53.6 & 47.2 & 50.8 & 45.4 & 49.4 & 54.0 & 51.0 \\
+  MSE & 58.4 & \textbf{61.2} & 52.4 & \textbf{61.4} & 50.4 & 56.8 & 55.4 & 55.2 & 55.6 & 44.4 \\
+ New Tok +  MSE & 58.8 & 58.0 & 54.4 & 58.6 & 48.6 & 56.4 & 51.0 & 53.6 & 55.8 & 45.6 \\
+  MRL & 55.8 & 59.8 & 51.6 & 57.2 & 51.0 & 55.6 & 53.6 & 52.2 & 51.8 & 50.6 \\
+ New Tok +  MRL & 55.2 & 59.2 & 52.0 & 55.8 & \textbf{52.0} & 57.0 & 51.2 & 53.2 & 51.8 & 50.6 \\
+ New Tok +  Multi & \textbf{60.0} & 57.8 & 53.8 & 60.6 & 50.0 & \textbf{59.6} & \textbf{55.6} & \textbf{56.0} & 55.4 & 47.0 \\
+ New Tok +  Multi + Aug & 59.0 & 60.2 & \textbf{54.6} & 61.2 & 50.8 & 58.0 & 53.8 & 54.4 & \textbf{57.6} & 45.6 \\
\bottomrule
\end{tabular}%
}
\caption{Finetuning on ACED---Preference accuracy on internal data, by language pair.}
\label{tab:results-pref_acc-segment-history-test-prefs-aced}
\end{table*}

\begin{table*}[htb]
\centering
\small
\resizebox{\textwidth}{!}{%
\begin{tabular}{lrrrrrrrrrr}
\toprule
\textbf{Model} & \textbf{en-ar} & \textbf{en-de} & \textbf{en-es} & \textbf{en-fi} & \textbf{en-fr} & \textbf{en-ja} & \textbf{en-pl} & \textbf{en-pt} & \textbf{en-sv} & \textbf{en-tr} \\
\midrule
\cometkiwibase & \textbf{0.175} & 0.119 & 0.103 & 0.133 & 0.173 & \textbf{0.164} & 0.136 & 0.086 & 0.110 & \textbf{0.235} \\
+ New Tok & 0.175 & 0.130 & 0.095 & 0.127 & 0.156 & 0.160 & \textbf{0.159} & 0.098 & 0.107 & 0.219 \\
+  MSE & 0.155 & \textbf{0.137} & \textbf{0.132} & \textbf{0.152} & \textbf{0.225} & 0.133 & 0.096 & 0.110 & \textbf{0.134} & 0.214 \\
+ New Tok +  MSE & 0.151 & \textbf{0.137} & \textbf{0.132} & 0.132 & 0.215 & 0.140 & 0.102 & 0.116 & 0.130 & 0.208 \\
+ MRL & 0.069 & 0.036 & 0.020 & 0.102 & 0.074 & 0.137 & 0.057 & -0.000 & 0.004 & 0.149 \\
+ New Tok +  MRL & 0.023 & 0.028 & -0.003 & 0.070 & 0.018 & 0.122 & 0.027 & -0.062 & -0.041 & 0.128 \\
+ New Tok +  Multi & 0.128 & 0.089 & 0.100 & 0.170 & 0.178 & 0.124 & 0.099 & 0.130 & 0.081 & 0.223 \\
+ New Tok +  Multi + Aug & 0.121 & 0.096 & 0.101 & 0.159 & 0.120 & 0.130 & 0.111 & \textbf{0.147} & 0.101 & 0.214 \\
\bottomrule
\end{tabular}%
}
\caption{Finetuning on internal data---Global Kendall's $\tau$ of QE models with human annotation z-scores, by language pair.}
\label{tab:results-kendall_corr-snap_internal}
\end{table*}

\begin{table*}[htb]
\centering
\small
\resizebox{\textwidth}{!}{%
\begin{tabular}{lrrrrrrrrrr}
\toprule
\textbf{Model} & \textbf{en-ar} & \textbf{en-de} & \textbf{en-es} & \textbf{en-fi} & \textbf{en-fr} & \textbf{en-ja} & \textbf{en-pl} & \textbf{en-pt} & \textbf{en-sv} & \textbf{en-tr} \\
\midrule
\cometkiwibase & 0.175 & 0.119 & 0.103 & 0.133 & 0.173 & 0.164 & 0.136 & 0.086 & 0.110 & 0.235 \\
+ New Tok & 0.175 & 0.130 & 0.095 & 0.127 & 0.156 & 0.160 & \textbf{0.159} & 0.098 & 0.107 & 0.219 \\
+  MSE & 0.133 & 0.113 & 0.086 & 0.110 & 0.135 & 0.148 & 0.080 & 0.037 & 0.123 & 0.228 \\
+ New Tok +  MSE & 0.145 & 0.138 & 0.111 & 0.139 & 0.131 & 0.158 & 0.095 & 0.076 & \textbf{0.133} & 0.253 \\
+  MRL & \textbf{0.186} & \textbf{0.133} & 0.123 & \textbf{0.162} & \textbf{0.253} & 0.192 & 0.128 & 0.122 & 0.101 & 0.257 \\
+ New Tok +  MRL & 0.180 & 0.131 & \textbf{0.124} & 0.148 & 0.221 & \textbf{0.193} & 0.143 &\textbf{ 0.128} & 0.107 & 0.249 \\
+ New Tok +  Multi & 0.152 & 0.105 & 0.098 & 0.133 & 0.178 & 0.147 & 0.097 & 0.044 & 0.088 & \textbf{0.279} \\
+ New Tok +  Multi + Aug & 0.162 & \textbf{0.133} & 0.112 & 0.151 & 0.182 & 0.157 & 0.126 & 0.089 & 0.127 & 0.254 \\
\bottomrule
\end{tabular}%
}
\caption{Finetuning on ACED---Global Kendall's $\tau$ of QE models with human annotation z-scores, by language pair.}
\label{tab:results-kendall_corr-snap_aced}
\end{table*}

\end{document}